\documentclass[acmtog,screen,nonacm]{acmart}

\usepackage{wrapfig}

\usepackage{eccvabbrv}

\usepackage{graphicx}
\usepackage{booktabs}
\usepackage{gensymb}
\newcommand{\tildeNice}{{\raise.17ex\hbox{$\scriptstyle\sim$}}}

\usepackage{hyperref}

\usepackage{orcidlink}

\usepackage{multibib}
\newcites{Main}{References}
\newcites{Supp}{Supplementary References}

\AtBeginDocument{%
  }

\begin{document}

\title{Radiance Fields from Photons}

\author{Sacha Jungerman}
\email{sjungerman@wisc.edu}
\orcid{0009-0008-4402-6444}
\author{Aryan Garg}
\email{agarg54@wisc.edu}
\orcid{0009-0003-7495-1060}
\author{Mohit Gupta}
\email{mgupta37@wisc.edu}
\orcid{0000-0002-2323-7700}
\affiliation{%
  \institution{University of Wisconsin-Madison}
  \city{Madison}
  \state{WI}
  \country{USA}
}
\setcopyright{cc}
\setcctype{by}
\acmJournal{TOG}
\acmYear{2025} \acmVolume{1} \acmNumber{1} \acmArticle{1} \acmMonth{1} \acmPrice{}\acmDOI{10.1145/3770578}

\begin{abstract}
    Neural radiance fields, or NeRFs, have become the de facto approach for high-quality view synthesis from a collection of images captured from multiple viewpoints. However, many issues remain when capturing images in-the-wild under challenging conditions, such as in low light, high dynamic range, or with rapid motion, leading to smeared reconstructions with noticeable artifacts. In this work, we introduce \emph{quanta radiance fields}, a novel class of neural radiance fields that are trained at the granularity of individual photons using single-photon cameras (SPCs). We develop theory and practical computational techniques for building radiance fields and estimating dense camera poses from unconventional, stochastic, and high-speed binary frame sequences captured by SPCs. We demonstrate, both via simulations and a SPC hardware prototype, high-fidelity reconstructions under high-speed motion, in low light, and for extreme dynamic range settings.  
\end{abstract}



\begin{teaserfigure}  %
  \centering
  \vspace{1em}
  \includegraphics[width=1.0\textwidth]{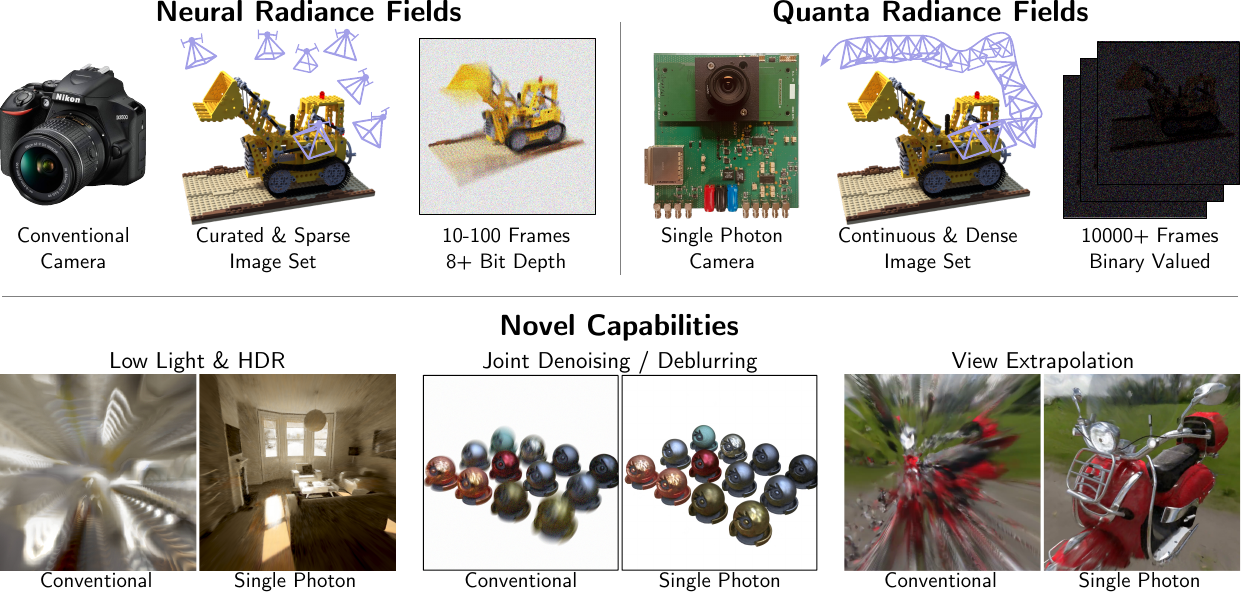}
  \vspace{-0.5em}
  \caption{
  \textbf{Single Photon Radiance Fields:} We introduce Quanta Radiance Fields (QRFs), neural radiance fields trained at the granularity of photons, using single photon cameras.  QRFs  significantly mitigate common challenges of conventional NeRFs and enable fast and continuous capture of the scene. They faithfully reconstruct scenes in extremely low-light and high dynamic range settings, effectively denoise and deblur training data without resorting to specialized techniques, and generalize better, producing novel view synthesis for a greater diversity of poses. \label{fig:teaser}
  }
\vspace{1.5em}
\end{teaserfigure}

\maketitle

\renewcommand*{\thefootnote}{$\ddagger$}
\setcounter{footnote}{1}
\footnotetext{This research was supported by the National Science Foundation via CAREER Award \#1943149, the Office of Naval Research via grant N000142412155, and the Wisconsin Alumni Research Foundation via a Research Forward Initiative Award.}
\renewcommand*{\thefootnote}{\arabic{footnote}}
\setcounter{footnote}{0}

\section{Introduction}
\label{sec:intro}

Whether they are used for autonomous navigation, localization, or augmented and mixed reality, a cornerstone of spatially intelligent systems is the ability to represent the world around us. Neural radiance fields~\citeMain{mildenhall2020nerf}, or simply NeRFs, have recently become an attractive choice for scene representations as they capture both appearance and geometry. 
NeRFs fundamentally operate on a set of pixel intensity measurements that are back-projected into a neural volume. From there, volumetric radiance models learn a scene representation in the form of a view-dependent pointwise color and opacity function, which produces the input pixel values when integrated along light rays corresponding to that pixel.

Under favorable imaging conditions, NeRFs can be built from a set of pixel measurements (collection of images) captured using conventional cameras and can enable high-fidelity scene reconstructions. 
However, in real-world scenarios, pixel intensities often suffer from artifacts such as motion or optical blur, strong noise in low-light settings, saturation in high-dynamic range scenes, and non-linearities due to proprietary image sensor processing pipelines. 
State-of-the-art NeRF techniques suffer dramatically -- or even fail entirely -- when the pixel data they consume contains such real-world imperfections. 
These issues are well-known and sufficiently important to have brought about long lines of work that aim to address each of these shortcomings individually~\citeMain{rawnerf,pearl2022nan,lee2023exblurf,Deblur-NeRF,Lee_2023_CVPR_DPNeRF}. 

We argue that many of these problems stem from the use of \emph{pixels as the atomic measurement unit} of visual information. 
Since the imaging artifacts (noise, blur, saturation, non-linearities) occur when pixel values are captured, these imperfections get ``baked-in'' the learned radiance fields, making it extremely challenging, if not impossible, to disentangle or mitigate them after the fact.

\smallskip
\noindent\textbf{Building Radiance Fields, One Photon at a Time:} Can we take a more granular approach, and build scene representations at the granularity of individual photons -- the finest scale at which visual information can be captured? If we had access to every photon in a scene, then, by definition, we would have captured the perfect radiance field, avoiding the artifacts mentioned above. 
Such Quanta Radiance Fields (QRF) -- radiance fields built one photon at a time -- would faithfully capture the photometric information in the scene, from complex specularities to varying albedoes and intricate geometry. Fortunately, there is an emerging class of single-photon cameras that are capable of detecting and counting individual photons~\citeMain{FossumQIS, SwissSPAD2} at ultra-high speeds, reaching up to $100$ kHz. These cameras are starting to become widely available, including in recent consumer devices (e.g., Apple iPhones), making them ideally suited to capture quanta radiance fields. 

As seen in Fig.~\ref{fig:teaser}, by using photons as the granular unit of visual information, QRFs considerably mitigate many of the common problems that plague traditional neural radiance fields, achieving high-quality reconstructions even under extreme imaging conditions such as large motion blur or strong camera noise in low-light and high dynamic range scenes. This is most notable in extremely low flux settings where conventional NeRF reconstructions are washed out due to the sensor's read-noise being baked into the neural volume. Furthermore, for a given total capture time, high-speed single-photon cameras sample a denser set of viewpoints, resulting in higher fidelity scene geometry estimation. This greater generalizability, referred to as view extrapolation in Fig.~\ref{fig:teaser}, enables novel view synthesis for views that are far from the training data. In contrast, these added viewpoints are integrated out by conventional cameras due to their lower frame rates leading to the characteristic cloudy or ghost-like artifacts seen in some NeRFs which are a symptom of poorly constrained geometry. Finally, with single photon cameras, one can continuously sample the scene as the camera moves through space. The resulting QRF can use the entire data sequence as input, without needing careful curation. Practically, this means that the training data can be captured considerably faster and more seamlessly, not only due to the high-speed nature of single-photon cameras but also because the user does not need to carefully plan or pause to take sharp images.

\smallskip
\noindent\textbf{Why is it Challenging to Build QRFs?} Although QRFs promise unprecedented scene representation capabilities, creating QRFs presents a unique set of challenges due to the unconventional image formation model of single-photon cameras, which capture photons as a high-speed sequence of binary frames: a pixel is ``on" if at least one photon is detected during the exposure time and ``off" otherwise. Many algorithms on which neural representations rely, such as feature matching, photometric pose optimization, and volume rendering, are not directly compatible with individual binary frames which suffer from severe noise and are not directly differentiable due to their discrete (binary) nature. One could integrate long sequences of binary frames over time to lower noise and quantization, but this comes at the cost of large motion blur, thus leading to a noise-vs-blur tradeoff. Our main observation is that it is possible to simultaneously avoid both blur and noise in QRFs by dense single-photon camera pose optimization, which allows aggregating information from a large collection of binary frames directly within the neural volume. We design a novel pose optimization regularizer tailored for high-speed single-photon cameras that enables poses corresponding to hundreds of thousands of frames to be learned simultaneously. 

Another considerable challenge is that of data deluge: While QRFs are trained for the same total number of optimization steps as their traditional counterparts, their dataset is made up of $10$s of thousands of noisy binary frames, as compared to a few $10$s of images in traditional NeRF methods. This data volume can easily overwhelm even high-end GPUs and, even if the hardware could keep up, it can lead to training times that scale with the number of input frames, which would render QRFs completely impractical. To mitigate these issues, we devise novel dataloading schemes to handle massive amounts of data captured by single-photon cameras allowing sublinear growth of training times. In practice, this scheme allows us to train QRFs with only about a $20\%$ overhead as compared to state-of-the-art radiance field methods despite using multiple orders of magnitude more frames, making it practical to build a representation that uses individual photons as basic building blocks.

\smallskip
\noindent\textbf{Scope and Limitations:} In this work, we take the first steps towards demonstrating that building scene representations at the granularity of individual photons enables high-fidelity view synthesis and 3D reconstructions in extremely challenging scenarios that were hitherto considered impossible. Even under normal conditions, we show that quanta radiance fields enable better reconstruction and view extrapolation as compared to their conventional counterparts. We show results on a wide range of imaging scenarios using both simulations and real captures using our prototype single-photon camera. Finally, we extend these ideas for use with the Gaussian splatting framework.

Thinking about radiance fields at the photon level may simultaneously address many of the common problems faced by neural radiance representations, however, many challenges remain. While single-photon cameras are becoming more common, this technology is not yet fully mature. Notably, current-generation sensors have limited resolution, lack color filter arrays\footnote{The color results shown here are in simulation.}, and incur high memory requirements. Further, although the proposed pose optimization scheme improves the reconstruction quality, poses still need to be initialized using conventional techniques. In some settings, this could become a limiting factor as conventional structure-from-motion techniques may fail before single-photon sensing. Addressing these limitations is necessary before QRFs can be widely adopted, and therefore are important next steps.

\section{Related Work}
\label{sec:related-work}

\noindent\textbf{Reconstruction with Single Photon Cameras:} While many technologies exist that enable detecting individual photons~\citeMain{maPhotonnumberresolvingMegapixelImage2017}, cameras based on single photon avalanche diodes (SPADs) technology are becoming prevalent due to ease of manufacturing, low cost, and high-speed capture. 

These sensors are most often used with active illumination, enabling them to directly measure depth via time-of-flight. SPAD-enabled solid-state LiDARs are widely deployed in automotive applications and have been used for $3$D reconstruction tasks~\citeMain{malik2023transient,malik2024flying,luo2024transientangelo,Mu24Towards3DVision} and non-line-of-sight imaging~\citeMain{Faccio_2020,nlos-km}. Alternatively, SPADs can be used entirely passively without a controlled light source much like a typical camera. 

In fact, passive single-photon cameras make excellent general-purpose imagers, having a large dynamic range~\citeMain{Ingle_CVPR2019_highflux,Liu_2022_WACV_extremehdr}, enabling fast motion compensation and reconstruction~\citeMain{qbp,Jungerman_2023_ICCV}, ultra-wideband imaging~\citeMain{wei2023ultrawideband}, and low-light inference~\citeMain{Goyal_2021_ICCV}. 
However, unlike their active counterparts, they cannot directly sense depth, making $3$D reconstruction significantly more challenging. In this work, we introduce a method for $3$D reconstructions and novel view synthesis for \textit{passive} SPAD-based single-photon cameras.

\smallskip \noindent \textbf{Burst Denoising Approaches:} Many works have focused on burst denoising of images~\citeMain{Hasinoff2016, chen2018learningdark, mildenhall2018burstdenoisingkernelprediction,lecouat2022highdynamicrangesuperresolution, nightsight}. 
These works have also inspired recent single-photon specific burst approaches~\citeMain{qbp,Jungerman_2023_ICCV,seets_2021_wacv,iwabuchi2021}.
Most of these methods are based on estimating and correcting for local motion in the image space. 

This approach, while effective under certain scenarios, is prone to alignment errors under challenging conditions (e.g., high noise, motion). These errors compound as any misalignment in pre-processing would introduce artifacts down the line. We propose a different approach of directly estimating this motion as part of the $3$D reconstruction process. 
This is similar in spirit to RawNeRF~\citeMain{rawnerf}, which found that NeRFs can act as general-purpose denoisers, far surpassing the denoising capabilities of a two-step approach. 

Further, conventional burst photography methods rely on optical flow to estimate and compensate for scene motion, which is compute-intensive, especially at high SPC framerates. 
For example, QBP~\citeMain{qbp}, a single-photon \textit{burst} photography technique, takes about $30$ minutes to merge a few thousand SPC frames into a \emph{single} denoised image. 
In contrast, our method can fully train on the entire scene with minimal artifacts in the same time QBP needs for a \emph{single} 3D reconstruction.

\smallskip\noindent\textbf{Neural Radiance Fields:} NeRFs~\citeMain{mildenhall2020nerf} enable view synthesis by modeling the scene as a neural network, implicitly baking in lighting and albedo. The scene is represented implicitly and estimated by minimizing the photometric error between the observed data and a rendering of the learned scene. Many subsequent works have addressed the original shortcomings of this method, by improving its speed with spatial datastructures~\citeMain{yu2021plenoctrees,mueller2022instant}, its original reliance on external pose estimates~\citeMain{lin2021barf}, or even issues regarding aliasing~\citeMain{barron2021mipnerf, barron2023zipnerf}. All these works consider a pixel as the atom of the visual representation -- or more precisely, the ray or field-of-view corresponding to each pixel during its exposure time -- whereas we propose using a finer unit of visual information, the photon, and learn not the scene radiance, but the view-dependent photon detection probability.

\smallskip\noindent\textbf{Radiance Fields for Challenging Scenes:} Creating radiance fields \emph{in-the-wild} under nonfavorable imaging conditions remains challenging. Sensor noise and motion blur can result in poor reconstructions with a characteristic cloud-like appearance. Fast motion, low light, or high dynamic range can also significantly degrade reconstruction. Many methods have been developed to address these issues, although typically in a piecemeal manner. For example, a recent approach~\citeMain{rawnerf} trains NeRFs directly on the raw sensor data, improving low-light performance. Some methods focus on denoising by using learned priors to denoise an image sequence, treating radiance fields as a burst photography approach~\citeMain{pearl2022nan}. There are methods dedicated to deblurring, which work either by modeling motion blur as part of the rendering step~\citeMain{lee2023exblurf}, or by using deformable kernels to correct for different types of blur~\citeMain{Deblur-NeRF,Lee_2023_CVPR_DPNeRF}. Although these methods might be orthogonal, it is not clear whether they are compatible with each other and whether they could be combined to handle multiple artifacts. Our goal is to demonstrate that by building scene representations at the finest granularity that physics allows, QRFs can mitigate multiple challenging cases simultaneously.

\smallskip\noindent\textbf{3D Gaussian Splatting:}
Unlike NeRFs, 3D Gaussian Splatting~\citeMain{kerbl3Dgaussians} explicitly represents scenes with a collection of 3D Gaussians, bypassing the need for a deep network. 
These Gaussians are initialized from a point cloud generated by COLMAP~\citeMain{colmap}, and learnable properties like opacity and spherical harmonic coefficients~\citeMain{plenoxels} for colors are optimized by minimizing photometric and structural errors between the observed data and the rendered scene. 
Subsequent works have addressed key limitations of the original method by eliminating the need for external pose estimates by using a pretrained monocular depth estimator~\citeMain{colmap_free_3dgs}, and efficiently managing memory scaling as point density increases~\citeMain{girish2024eaglesefficientaccelerated3d}. Others have focused on challenging conditions, such as handling motion blur~\citeMain{deblurring_gaussians} and improving performance in HDR and low-light scenarios~\citeMain{hdrsplat}.
While we preliminarily explore using splatting with single photon data in section~\ref{sec:splat}, this is an interesting avenue for future research.

\section{Neural Radiance Fields: Background}

\begin{figure*}[t!]
    \centering
    \includegraphics[width=1.01\textwidth]{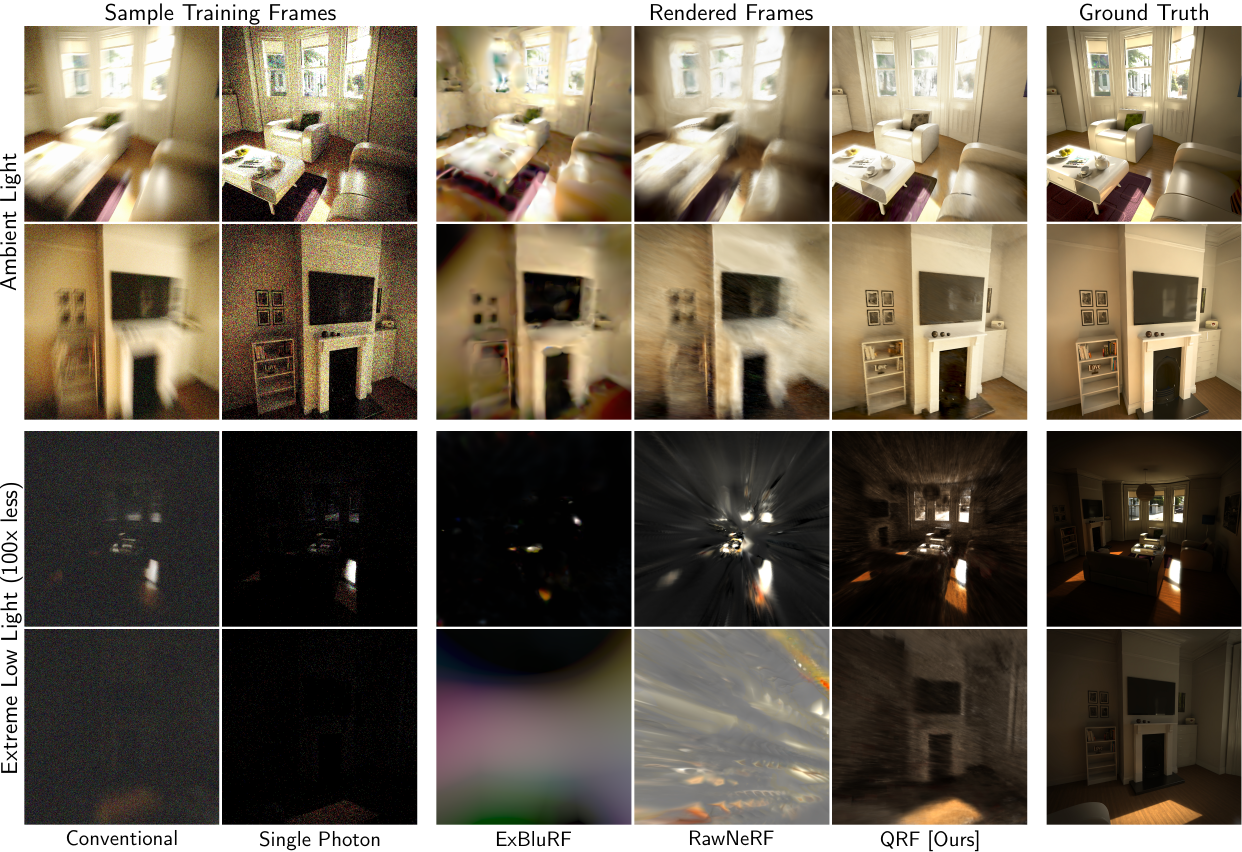}
    \vspace{-2em}
    \caption{\textbf{Reconstruction under challenging scenario:} We simulate a high-speed ($\tildeNice 72$km/h) drone fly-through of an indoor scene which has $16$ stops of dynamic range. We train a conventional motion-aware NeRF with tonemapped images~\protect\citeMain{lee2023exblurf}, one with raw linear intensity conventional images like in~\protect\citeMain{rawnerf}, and a QRF with simulated binary frames from this scene. In ambient light, the ExBluRF and RawNeRF reconstructions suffer from various artifacts, while the QRF model captures the full dynamic range of the scene with no noticeable blur. In the extreme case of lowering the light levels by $100\times$, the QRF reconstruction remains recognizable, while baselines fail.
    \label{fig:lowflux-hdr}}
\end{figure*}

NeRFs learn the scene's radiance, $\mathbf{L}(x, d)$, and volume density, $\sigma(x)$, for any point in space by inverting the process by which the scene's radiance gets mapped to a camera pixel measurement. However, instead of integrating the radiance over the solid angle subtended by each pixel, NeRFs often approximate this by taking a volumetric ray-tracing approach. This simplifies the forward model, although it has been shown to cause aliasing issues that can be solved by integrating over the whole field-of-view~\citeMain{barron2021mipnerf,barron2023zipnerf}. Conventionally, the volume rendering equation used to render the expected radiant flux $\hat\phi(\mathbf{r})$ incident upon a pixel parameterized by a camera ray $\mathbf{r}(t)=\mathbf{o}+t\mathbf{d}$ with near and far bounds $t_n$ and $t_f$ is defined as~\citeMain{max1995_volume_rendering,mildenhall2020nerf,volumedensities}:

\begin{align}
\begin{split}
\hat\phi(\mathbf{r})=\int_{t_n}^{t_f} T(t) \sigma(\mathbf{r}(t)) \mathbf{L}(\mathbf{r}(t), \mathbf{d}) d t,\\
\text { where } T(t)=\exp \left(-\int_{t_n}^t \sigma(\mathbf{r}(s)) d s\right)
\label{eq:rendering}
\end{split}
\end{align}

No closed-form solution for $(\sigma, \mathbf{L})$ exists, so these are estimated using SGD from a set of ideal measurements $\phi_i$ and their poses. \\

\noindent\textbf{Learning Radiance Fields from Pixels:} The pixel brightness $I$ of a conventional CMOS camera can be modeled as a function of scene radiance, consisting of two simple transformations~\citeMain{GrossbergNayar_PAMI04}. First, light passes through the camera's optical stack, which may present aberrations and various defects. While incorrect focusing or issues with shallow depth-of-field can occur and have been the topic of recent works~\citeMain{Deblur-NeRF,Lee_2023_CVPR_DPNeRF}, we assume this transformation is linear, or equivalently that we use an in-focus camera with aberration-free optics. The image irradiance then hits the sensor and is converted to image brightness via a complex, nonlinear function $f$ that encapsulates the camera response curve, tone mapping, and proprietary image signal processing. In summary, we have $I = f(\phi)$.

The specific camera response function $f$ can vary from manufacturer to manufacturer, yet most CMOS pixels will eventually saturate as they reach their full well capacity (FWC), leading to limited dynamic range. On top of this, many sources of noise, denoted by $\mathcal{N}$, exist which affect the sensor's low-light performance. A simple model for this response function can be written as\footnote{For simplicity, various sources of noise, including photon noise, Gaussian sensor read noise, fixed pattern, and quantization noise, are absorbed into~$\mathcal{N}$.}:

\vspace{0.75em}
\begin{equation}
    I = f(\hat{\phi}) = min\left(\int_{\tau} \hat{\phi} dt,~\text{FWC}\right) + \mathcal{N} \,.
    \label{eq:camera_model}
\end{equation}
\vspace{1em}

where $\tau$ is the total exposure time. 
During training, the rendering equation Eq.~\ref{eq:rendering} is computed numerically by sampling points along a ray, and $(\sigma, \mathbf{L})$ are learned by minimizing a photometric loss between the rendered pixel color $\hat{\phi}$ and the observed pixel colors $I$:

\begin{equation}
    \mathcal{L}_{\text{photo}} = ||\hat{\phi} - I||^2_2 \,.
    \label{eq:std-photometric-loss}
\end{equation}

In so doing, a typical NeRF does not learn to recover the radiance of the scene $\mathbf{L}(x, d)$, since the above loss minimizes the difference between $\hat{\phi}$ and $I$ instead of $f(\hat{\phi})$ and $I$. That is, the nonlinear effects of the camera response function, pixel saturation, tone mapping, and noise, get baked in, as they are captured by the pixel intensities. Although using raw linear intensity pixels can help~\citeMain{rawnerf}, many of these issues, notably saturation, blur, and noise, remain.\\

\section{Building Radiance Fields, One Photon at a Time}
\label{sec:main_method}

Building radiance fields from conventional pixels is limiting in three fundamental ways. First, in low flux conditions, potentially severe noise $\mathcal{N}$ gets baked into the learned radiance field, washing out any reconstructions. This ``white noise'' phenomenon can be observed in the last row of Fig.~\ref{fig:lowflux-hdr}. Second, in high-flux settings, the pixels saturate, leading to poor contrast or clipped regions. Lastly, under rapid motion or long integration times $\tau$, pixel measurements may have large motion blur. 

In this section, we develop the theoretical foundations and practical methods for estimating quanta radiance fields, i.e., radiance fields built one photon at a time, and discuss how QRFs could address the limitations described above. 

\subsection{Quanta Sensors: Background} 
We start by describing the image formation model of SPAD-based single-photon cameras that we use to capture QRFs. SPADs are digital photon counting devices, and as such they do not suffer from read noise, making them only fundamentally shot noise limited~\citeMain{SPADnoiseAnalysis}. These capabilities enable high low-light sensitivity, high temporal resolution, and extremely large dynamic range.

\begin{figure*}[t!]
    \centering
    \includegraphics[width=1.0\textwidth]{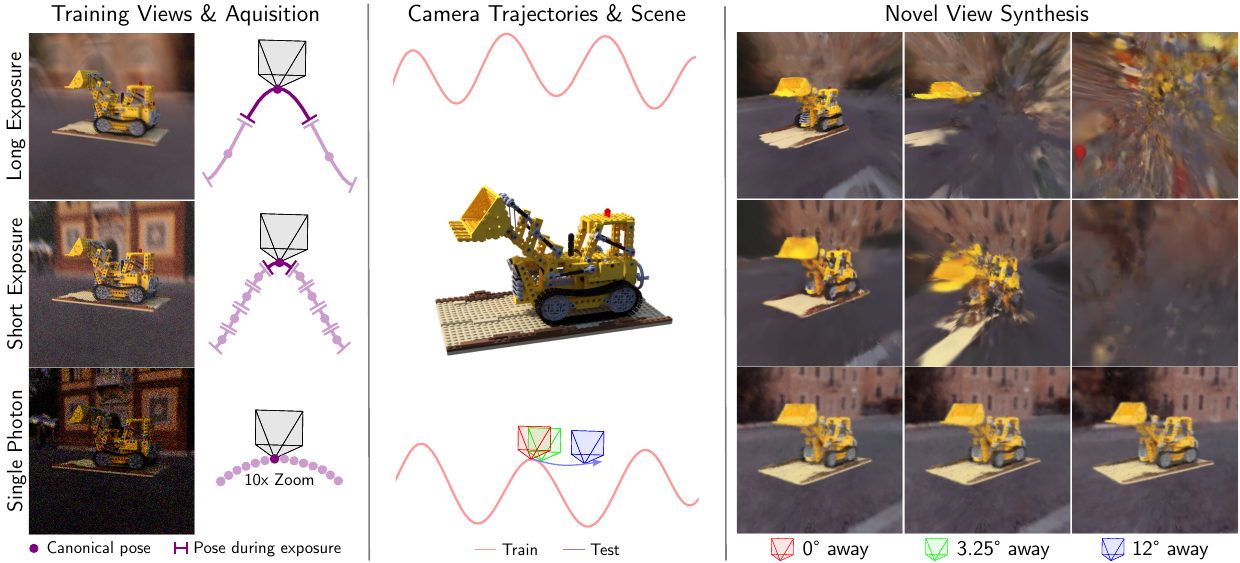}
    \caption{\textbf{View Diversity and Extrapolation:} 
    Radiance fields trained using single photon data perform better view extrapolation (novel views that are significantly different than the training poses) and degrade more gracefully than ones trained with conventional frames, given the same total capture time. The training viewpoints in both cases span the same trajectory, that is, both datasets have the same pose diversity, however, the denser sampling provided by the single photon camera better constrains the scene's geometry leading to significantly improved reconstructions and better generalization. 
    \label{fig:view-interp}}
\end{figure*}

Consider a SPAD pixel observing a scene with a radiant flux of $\phi$. The number of incident photons $k$ on a pixel during an exposure time $\tau$ follows a Poisson distribution given by:

\begin{equation}
    P(k) = \frac{(\phi\tau)^ke^{-\phi\tau}}{k!} \,.
\end{equation}

However, a SPAD pixel resets after each photon detection. During this ``dead'' time, the pixel cannot detect any more photons. Thus, the measurement $B$, of a SPAD pixel is binary ($1$ if the pixel records one or more photons during the exposure time $\tau$, $0$ otherwise) and follows a Bernoulli distribution given by:

\begin{equation}
\begin{split}
    P(B=0) &= P(k=0) = e^{-\phi \tau},\\
    P(B=1) &= P(k \ge 1)=1-e^{-\phi \tau}.
    \label{eq:bernoulli}
\end{split}
\end{equation}

Notice how this imaging model is different from the one described by Eq.~\ref{eq:camera_model}. There is no read-noise or full well capacity and $\tau$ is usually in the tenths of microseconds range as opposed to the tenths of millisecond range. 

\smallskip\noindent\textbf{Two Step Reconstruction:} The key challenge in estimating the radiance field from quanta sensors is the highly quantized and noisy nature of their raw binary measurements. One idea is to preprocess the input time series and use these to learn a radiance field. 

A minimal preprocessing step is to add consecutive binary frames together and generate ``virtual exposures''~\citeMain{Jungerman_2023_ICCV} to mitigate noise and quantization. However, for dynamic scenes, this approach runs into the fundamental noise-vs.-blur trade-off. Similarly to a conventional camera image, if the total exposure time $n\tau$ of a virtual exposure produced by aggregating a sequence of $n$ binary frames is large compared to the motion of the camera or scene, the resulting virtual exposure image will be blurred. Unlike a conventional image, however, the parameter $n$ can be reduced post-capture by changing the exposure time after the fact, trading off motion blur and noise.

Instead of settling for an operating point on this blur-versus-noise trade-off space, many techniques have proposed motion-adaptive integration of binary frames, or burst processing approaches~\citeMain{qbp,Jungerman_2023_ICCV,seets_2021_wacv,iwabuchi2021}. However, two main issues arise from using these: i) any misalignments or preprocessing artifacts cannot be fixed during $3$D reconstruction, leading to compounding errors, and ii) these methods can be extremely compute-intensive, with QBP~\citeMain{qbp} taking about $30$ minutes to merge a few thousand binary frames into a \emph{single} denoised image.

We instead aggregate measurements directly in the neural volume, allowing us to bypass the expensive preprocessing step -- enabling QRFs to learn an entire scene in the same time QBP takes to create a single denoised image -- and re-cast the noise-vs-blur tradeoff as a constrained pose optimization and $3$D reconstruction problem.

\smallskip\noindent\textbf{Pose Diversity vs. Sampling rate:} Given the same camera trajectory and time budget, a single-photon camera will capture many more frames than a conventional camera due to its high sampling rate. However, both cameras will sweep through a range of poses during each of their respective exposure times, shown schematically in Fig.~\ref{fig:view-interp}~(left) as an interval centered around what we call the \textit{canonical pose}, which all contribute to the captured image.
The extent of this interval, and thus the resultant blur, varies predominantly based on the exposure time. In its limit, the interval becomes vanishingly small as the trajectory is sampled at higher rates. Fig.~\ref{fig:view-interp} shows simulated frames for a conventional RGB camera (at $50$, and $200$ fps) and a single photon camera (at $80$ kHz) alongside their poses and their support\footnote{The \textit{support} of each canonical pose also extends to camera rotations.}.

However, in all cases, the same \textit{pose diversity} is encountered\footnote{Assuming a large shutter angle throughout, which is crucial for a good light efficiency.} as both cameras sweep the same path. Crucially, the difference is that for conventional cameras, the motion is entangled into the captured frame as motion blur, whereas not with SPADs.

\subsection{Estimating Radiance from Quanta Measurements:} 
Typically, NeRFs are trained using images that have been processed on-device using a complex, and often proprietary, pipeline. Previous work~\citeMain{rawnerf} has shown that learning a radiance field using raw, mosaicked, linear intensity images and deferring post-processing to after-training results in cleaner denoised images and a larger dynamic range. Single photon cameras enable us to push this idea to its limit: instead of training using linear intensity images, which can be achieved using virtual exposures, we learn a radiance field directly from binary frames.

The induced change of domain on the learned radiance field means that the network learns the spatially varying, view-dependent, probability $\widehat{P}$ that at least one photon is detected. Similarly to previous works, what is learned by the network still isn't true radiance. However, we can trivially invert the SPAD's camera response (Eq.~\ref{eq:bernoulli}) and get a good estimate of scene radiance:

\begin{equation}
    \label{eq:flux-estimate}
    \hat{\phi}=-\frac{1}{\tau} \ln \left(1-\hat{P}\right)\,.
\end{equation}

Note that this is similar to the maximum likelihood estimator of $\phi$ for a static scene~\citeMain{oversampled}, except that here the photon detection probability $\hat{P}$ is estimated using Eq.~\ref{eq:rendering} by sampling the network along rays which move with the camera instead of by the average of many consecutive binary frames. Using Eq.~\ref{eq:flux-estimate}, we can effectively estimate the scene's flux and render tonemapped images. Learning the photon detection probability directly avoids having to average binary frames, which might introduce motion blur and numerical instabilities that can occur in Eq.~\ref{eq:flux-estimate}, and does not require any preprocessing of the raw binary data.

Thus, we train neural radiance fields directly on the binary measurements $B$ captured with an array of SPAD pixels by defining a photometric loss term on binary measurements:

\begin{equation}
    \mathcal{L}_{\text{quanta}} = ||\hat{P} - B||^2_2 \,.
    \label{eq:bin-photometric-loss}
\end{equation}

Where $\hat{P}$ is the rendered photon detection probability, which is computed using Eq.~\ref{eq:rendering}.
Although each measurement $B$ is extremely noisy and quantized, the continuous-valued photometric loss in Eq.~\ref{eq:bin-photometric-loss}, and continuously estimated photon detection probability, ensure the differentiability of the resulting optimization problem. The training process can be noisy, but stability is ensured thanks to the robustness to noise that modern optimizers provide. In addition, using a large batch size and learning weight decay helps increase stability.

\subsection{Pose Optimization with Quanta Cameras}
\label{sec:pose-opt}

In practice, poses are often initialized using estimates obtained from a structure-from-motion preprocessing step such as COLMAP~\citeMain{colmap} or from an IMU chip. As these initial estimates often suffer from noise and drift, camera poses need to be co-optimized alongside the radiance field. While most modern NeRF variants perform this optimization by default, it is not feasible when directly using binary frames. The problem is twofold: i) each individual frame is too noisy, leading to noisy optimized poses that are prone to get stuck in local minima, and ii) the number of poses that need to be optimized is greatly increased, from a few hundred to a few hundred thousand, which can make the optimization intractable without careful considerations.

\smallskip\noindent\textbf{Motivation \& Design:} Optimizing poses of conventional cameras is an already notoriously difficult and non-convex problem, which gets considerably harder when every image is binary valued as the photometric loss becomes noisy and unreliable. Our key enabling observation is that, due to the high sampling rate of single-photon cameras and known frame ordering, we can leverage a simple but powerful prior: neighboring frames should have similar poses. More formally, the ``pose trajectory" should be smooth. 

A common way to enforce smoothness is to optimize a lower-dimensional representation of the trajectory as opposed to the camera poses. Lee \etal~\citeMain{lee2023exblurf} optimize a B\'ezier curve for each view in order to learn the range of poses each camera sweeps through during its exposure, in other words, the support of each canonical pose. This over-parametrizes each pose by the order $M=7$ of the spline, with all new control points $\{{\hat{\mathbf{p}}}_{j}\}_{j=0}^{M}\in{\mathfrak{s e}}(3)$ being initialized to the original pose. This could be adapted to work with binary frames by, for instance, optimizing an $M^{\text{th}}$ order spline for groups of $N$ binary frames, where a natural choice of $N$ would be the ratio of frame rates between a single photon and conventional camera. While this would lower the learnable pose parameters by $M/N$, two main issues arise. First, it only considers smoothness within the window size $N$, and does not consider how these sub-splines piece together to form the whole trajectory. Second, initializing each subtrajectory would require solving for a best-fit spline, which quickly becomes unwieldy given the huge number of binary frames.  

Instead, we devise a Fourier-domain regularizer to formalize the smoothness insight and apply it to pose optimization. Although smoothing the trajectory in this manner does not directly lower the dimensionality of the camera trajectory, it sidesteps potentially expensive spline-fitting computations and remains extremely fast as we can leverage modern FFT implementations, which have been highly optimized and parallelized to run on GPU. However, the main benefit is that this allows us to think about camera trajectories and shake in a natural way, using frequencies.

Specifically, the trajectory can be filtered in the Fourier domain. For instance, a low-pass filter can effectively smooth out high frequencies which are mostly composed of noise. More complex filtering regimes are also possible, for instance, a notch or band-stop filter can prove useful for filtering out specific frequencies such as vehicle resonant frequencies for vehicle-mounted cameras or handshake jitter for handheld cameras.

Crucially, this frequency-based reasoning is enabled by the high sampling rate of SPADs, and is not easily applicable to conventional cameras. The latter do not sample the scene fast enough to capture many frequency components without aliasing. In fact, the proposed Fourier regularizer naturally collapses to zero when used with conventional cameras as the cutoff frequency of the low-pass filter becomes larger than the Nyquist rate of the camera.

\smallskip\noindent\textbf{Fourier Pose Regularizer:}
We first encode poses as $9$-dimensional vectors consisting of a translation component $\mathbf{t} = [x,y,z]^T$, and a rotational component~\citeMain{Zhou_2019_CVPR_Continuity_of_Rotation}. The rotational mapping is not unique, as it is over-parameterized, yet it is smooth, invertible, and easily computable. The resulting tensor, $\mathcal{P}$, has dimensions $N\times9$ where $N$ is the number of poses (number of binary frames in the captured sequence). 

These components are then individually smoothed using a low-pass filter in the Fourier domain, transformed back, and compared to their non-smoothed counterparts, resulting in the following total loss:
\begin{equation}
\begin{split}
    \widehat{\mathcal{P}} &= \mathcal{F}^{-1}(\mathcal{F}(\mathcal{P}) \cdot H_{\text{lowpass}}) \\
    \mathcal{L}_{\text{total}} &= \mathcal{L}_{\text{quanta}} + \lambda \sum_{j} ||\mathcal{P}_j - \widehat{\mathcal{P}}_j||^2
    \label{eq:fft-reg}
\end{split}
\end{equation}

\begin{figure*}[t!]
    \centering
    \includegraphics[width=1.0\textwidth]{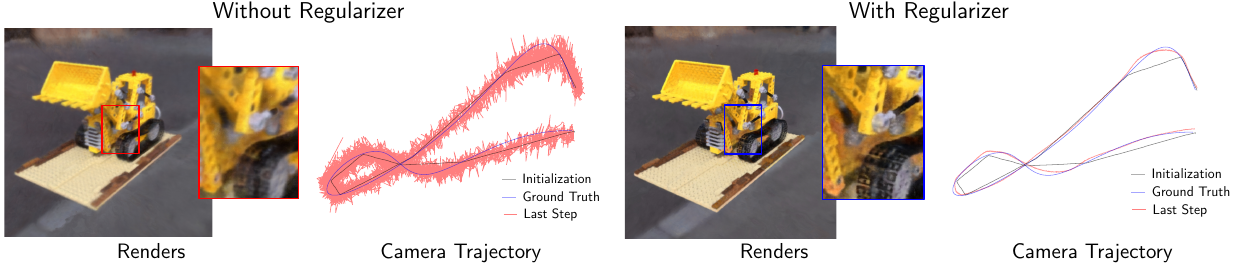}
    \vspace{-2em}
    \caption{\textbf{Camera Pose Optimization:} The trajectory of the camera is co-optimized with the radiance field. Due to the large number of camera poses to optimize and the noisy binary measurements, a strong smoothing regularizer on the poses is needed. Without it, poses settle in noisy local minima, which affects the final reconstruction quality. \label{fig:poseopt}}
\end{figure*}

Where $\lambda$ controls the regularizer strength, $\mathcal{F}$ is the $1$D Fourier transform (which is applied only along the first dimension of the pose embedding), and $H_{\text{lowpass}}$ is the transfer function of a lowpass filter. This smoothing is performed on the 9-dimensional vectors representing camera poses. As seen in Fig.~\ref{fig:poseopt}, a strong smoothing regularizer is imperative to obtain visually pleasing results when training NeRFs with high-speed SPC data. With the regularizer ensuring that the trajectory is well-behaved, the high-speed SPAD sampling can recover fine tremors and high-frequency motion.

\smallskip\noindent\textbf{Pose Embedding Validity:} The $6$D rotational component of the pose embedding consists of the rotation matrix's first two columns, namely $i, j$ concatenated together. To undo this mapping, we follow Eq.~16 in~\citeMain{Zhou_2019_CVPR_Continuity_of_Rotation}, which extracts these vectors, normalizes them, and ensures they are orthogonal, before computing the last column of the rotation matrix as their cross product. This mapping is invertible and results in a valid $SO(3)$ rotation matrix as long as the span of these two column vectors is two-dimensional.

While this is true of any initial poses, it might not hold for smoothed ones. Specifically, $i, j$ might either be colinear, or at least one of them might be the zero vector. It is also possible to redo the rotation embedding, which renormalizes these components and effectively snaps them back onto the $SO(3)$ manifold. While such degenerate cases are theoretically possible, in practice, we have not encountered these in our experiments.

\smallskip\noindent\textbf{Filter Design \& Computational Considerations:} While we use a low-pass filter in Eq.~\ref{eq:fft-reg}, many other filters could be employed. For our use case, there are two main factors to consider when picking a filter: trajectory boundaries and phase delay. 

First, if the camera trajectory cannot be assumed to be periodic, then careful consideration around the endpoints of the camera trajectory is needed. For best results, we follow conventional filtering practices and pad the signal before filtering. Specifically, we pad the pose trajectory on either end by a small amount using linear extrapolation, perform filtering, and then crop out the padded poses.     

Second, it is important to realize that, in many cases, filtering the trajectories can introduce a nonzero phase shift. As we iteratively minimize the distance between the current and smoothed trajectories, this phase shift can lead to the trajectory drifting over time. In all our experiments, we simply use an ideal low-pass filter (``brick wall'' filter), and despite having a linear phase response or equivalently a constant group delay, we empirically notice that with a small enough $\lambda$ this delay is not an issue as it is corrected by pose optimization. Furthermore, filtering does not have to be causal, meaning that we can correct for phase delays explicitly, or use filters that introduce little to no phase shift such as the Savitzky-Golay filter, which can be thought of as locally fitting a polynomial to the input signal, although there are better options~\citeMain{Schmid2022WhyAH}. 

Finally, while Fourier transforms are fast thanks to the FFT algorithm, repeatedly computing them at every step could become computationally expensive. This can be partially mitigated by using the Fourier pose regularizer every $n$ steps, and replacing $\lambda$ by $n\lambda$ in Eq.~\ref{eq:fft-reg}. Empirically, we found that this tradeoff resulted in a noisier loss, and did not contribute to significant wall-time savings in part due to the fast implementation of pytorch's \texttt{fft.rfft} method. Instead, the main bottleneck is dataloading which we tackle next.

\subsection{Data deluge \& Practical Considerations} 
The extremely high-speed capture enabled by single-photon cameras can easily strain the available bandwidth and memory of a system as a large quantity of data gets acquired rapidly. For example, a current SPAD-based single-photon camera~\citeMain{SwissSPAD2} has a modest resolution of $512\times512$ and can run at $100$kHz, resulting in a bandwidth of $24.4$ Gb/s, more than two orders of magnitude more than for a conventional camera with similar specs running at $60$ fps ($\sim0.1$ Gb/s). This data deluge problem has been the subject of many recent works~\citeMain{codedaperture,sundar2023sodacam,CompressiveSPC}, however, these usually compress the data in a lossy way and cannot be directly used in our context for building neural scene representations. 

Further, NeRFs are trained on mini-batches of pairs of rays and pixel values. The implication is that, at each step, a random sample of pixels must be drawn uniformly from all the training data. For conventional images, this is feasible since a few hundred images can be decoded and cached on GPU as one big tensor. However, this is infeasible for binary frames because of prohibitively large amounts of data, which pose an acute technical challenge. 

To solve both of these problems, we bit-pack the binary frames, which provides an $8\times$ compression, and memory-map the whole bit-packed array. Our dataloader is then responsible for loading binary pixel data directly from the disk, decompressing and extracting the individual bits on the fly, and sending them to the GPU. Despite training on potentially hundreds of thousands of frames, we find that, with modern solid-state drives, this data-loading scheme is only about $20\%$ slower than when training with around a hundred conventional images that are preloaded and cached on the GPU. However, when using slower conventional hard disks the training time can easily double. Bit-packing the array does not significantly impact training time, rather it makes the dataset's disk footprint more manageable and might contribute to better cache locality.

\begin{figure}[t!]
    \centering
    \includegraphics[width=1.0\columnwidth]{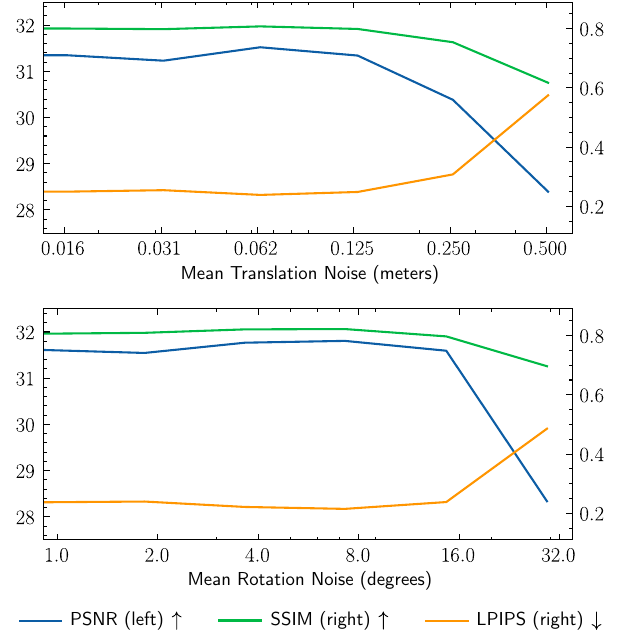}
    \vspace{-2em}
    \caption{\textbf{Robustness of Pose Regularizer:} Using the same scene as in Fig.~\ref{fig:poseopt}, we show the final reconstruction quality as a function of the amount of noise added to the pose initialization. Reconstruction quality stays nearly constant, rivaling that of using ground truth poses, up until it rapidly collapses when initialized with around $0.25$m and $16\degree$ of mean translational or rotational noise. For comparison, the base of the truck is about $2.3\text{m}\times1.3$m and the camera's diagonal field-of-view is $39.6\degree$.    
    \label{fig:poseopt-metrics}}
\end{figure}

\begin{figure*}[ht!]
    \centering
    \includegraphics[width=1.0\textwidth]{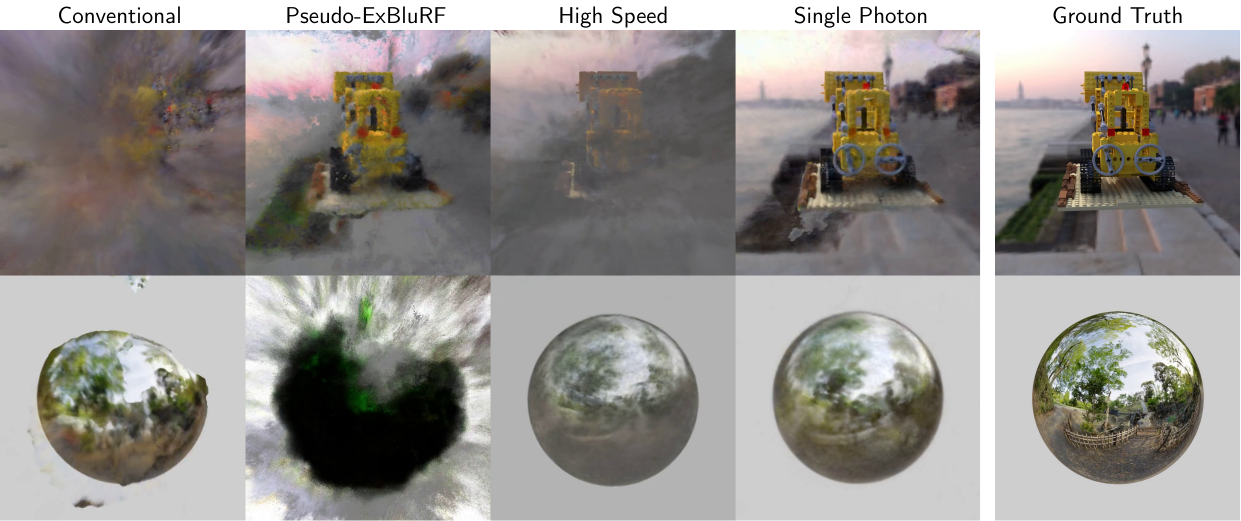}
    \caption{\textbf{Robustness to Motion Blur:} (Column 1) Traditional NeRF reconstructions suffer considerably when the data it has been trained on contains motion blur. (Column 2) Some methods, such as~\protect\citeMain{lee2023exblurf}, address this by incorporating the formation of image blur into the forward rendering model. (Column 3) A high-speed camera running at 1000 fps can be used to capture more views with less motion blur, however, the colors look washed out due to the read noise being baked into the learned radiance field. (Column 4) We show that, despite much noisier individual frames, training using binary data obtained from a single photon camera achieves visually superior reconstructions. For corresponding quantitative results, see Tab.~\ref{tab:view-interp}.\label{fig:blur}}
\end{figure*}

\section{Novel Capabilities of Quanta Radiance Fields}

\begin{table}[t!]
    \centering
    \caption{\textbf{Reconstruction under challenging scenario:} quantitative evaluation of scenes shown in Fig.~\ref{fig:lowflux-hdr}.}
    \vspace{-0.5em}
    \scalebox{1.1}{
        \begin{tabular}{l|ccc}
            \hline 
            \multicolumn{1}{c|}{Method}   & PSNR$\uparrow$      & SSIM$\uparrow$      & LPIPS$\downarrow$     \\ \hline \hline
            ExBluRF (Ambient)             & $16.35$             & $0.687$             & $0.500$              \\
            RawNeRF (Ambient)             & $10.95$             & $0.466$             & $0.308$               \\
            QRF (Ambient)                 & $\mathbf{25.90}$    & $\mathbf{0.779}$    & $\mathbf{0.247}$      \\
            \hline
            ExBluRF (Low Light)           & $15.83$             & $0.397$             & $0.357$               \\
            RawNeRF (Low Light)           & $\mathbf{17.78}$    & $0.681$             & $0.176$               \\
            QRF (Low Light)               & $17.12$             & $\mathbf{0.804}$    & $\mathbf{0.093}$      \\
            \hline
        \end{tabular}
    }
    \label{tab:denoising}
\end{table}

\begin{table}[t!]
    \centering
    \caption{\textbf{View Extrapolation:} quantitative evaluation of novel view synthesis averaged over all simulated scenes.}
    \vspace{-0.5em}
    \scalebox{1.1}{
        \begin{tabular}{l|ccc}
            \hline
            \multicolumn{1}{c|}{Method} & PSNR$\uparrow$      & SSIM$\uparrow$      & LPIPS$\downarrow$   \\ \hline \hline
            Conventional                & $12.58$             & $0.622$             & $0.165$             \\
            Pseudo-ExBLuRF              & $15.72$             & $0.493$             & $0.222$             \\
            High Speed                  & $\underline{17.35}$ & $\underline{0.700}$ & $\underline{0.108}$ \\
            QRF               & $\mathbf{19.74}$    & $\mathbf{0.752}$    & $\mathbf{0.094}$    \\
            \hline
        \end{tabular}
    }
    \label{tab:view-interp}
\end{table}

\noindent\textbf{Low Light and High Dynamic Range:} The excellent low-light and high dynamic range characteristics of single-photon cameras enable the creation of neural radiance fields in challenging scenarios that are impossible to capture with conventional cameras. In Fig.~\ref{fig:lowflux-hdr}, we simulate a conventional camera ($50$ fps) and a single photon camera ($10$ kHz SPAD) zipping through a scene with extremely high dynamic range ($\tildeNice 61,000$). We show ExBluRF~\citeMain{lee2023exblurf}, RawNeRF~\citeMain{rawnerf}, and QRF reconstructions, as well as raw frames, for both cameras. In both cases, the trajectory and total capture time are held constant. Here, we initialize the camera poses to their ground truth values and disable pose optimization\footnote{except for sub-exposures, which is needed for ExBluRF to account for motion blur.} to disentangle the effects of pose optimization from low-light performance.   

Under ambient light, the reconstruction trained using conventional camera frames struggles to properly reconstruct the scene due to camera motion and high dynamic range. Despite substantial noise in the single-photon input frames, the QRF reconstruction is high-fidelity, with sharp reflections on the hardwood floor and clearly discernible books on the shelf. 

Under extremely low light ($100\times$ lower than ambient light), the raw frames from both sensors are almost entirely dark, with fewer than $0.01$ photons per pixel detected by the single-photon camera. At these light levels, the read noise from the conventional camera completely overwhelms the training process, leading to distorted highlights at best and featureless gray reconstructions at worst (last row, conventional). In contrast, the single photon reconstruction degrades much more gracefully -- significant noise can be seen throughout, but the scene remains recognizable despite the extremely challenging conditions. Finally, Tab.~\ref{tab:denoising} shows quantitative evaluations for this scene.

\smallskip\noindent \textbf{Denoising and Deblurring:} NeRFs have been shown to be excellent general-purpose denoisers, beating even state-of-the-art one-shot denoisers, when accurate camera poses can be estimated~\citeMain{rawnerf}. With quanta cameras, we can take this idea to its physical limit, where the denoising and deblurring capabilities are only limited by the fundamental shot noise of photon arrival. 

We demonstrate these capabilities in Fig.~\ref{fig:lowflux-hdr}, where again, in all cases the total capture time and camera trajectories are held constant for a fair comparison. Already at medium light levels, the conventional reconstructions start to suffer from motion artifacts and blown-out highlights. With $100\times$ less light, reconstructions made using raw conventional camera frames are washed out due to the inherent noise in the measurements. This bias at low flux is not seen when using SPCs as they are only shot noise limited.

\smallskip\noindent\textbf{View Extrapolation:} While neural radiance fields excel at novel view synthesis under ideal conditions, imperfections -- such as camera noise or blur -- cause typical methods to fail when the desired novel view is not close to a training view. We show that by using frames from a single-photon camera, which are individually noisier but can be captured at faster frame rates, we can perform \emph{view extrapolation} and not merely view interpolation. 
Here, extrapolation differs from interpolation in that it enables rendering from viewpoints that lack a corresponding or nearby training pose. Most NeRF models, however, tend to fail when generating views from poses significantly outside the training trajectory.

In Fig.~\ref{fig:view-interp}, we learn a NeRF of a Lego truck with simulated frames for a conventional camera at different framerates ($50$ and $200$ fps) and a single photon camera (at $80$ kHz). In all cases, the total capture time and the camera's trajectory are held constant. The training poses are drawn from a sinusoidal trajectory that encircles the object and is sampled at regular intervals corresponding to the camera's framerate. Once trained, we render frames from a validation trajectory which starts on the training trajectory and slowly gets further along a circular arc centered on the truck. In all cases, the camera faces the truck enabling us to easily measure the displacement between the training and testing views in degrees. Here, $12\degree$ away means that the test view is exactly between two peaks in the training trajectory ($360\degree/15$) and displaced by about one unit (radius $=5$). Notice that with a small perturbation of the pose of only a few degrees, the reconstruction quality for the NeRF trained with conventional frames degrades rapidly, and completely fails thereafter. Quantitative view extrapolation results for all simulated scenes are shown in Tab.~\ref{tab:view-interp}.

\smallskip\noindent\textbf{Deblurring as Pose Optimization:} Despite potential issues caused by blurry images when optimizing for pose, for conventional cameras, motion blur and pose estimation are generally considered as two distinct issues. This is not the case for quanta radiance fields. While each individual binary frame can be assumed to have no inherent motion blur due to its extremely small exposure time, poor pose estimates will cause the learned radiance to appear blurry, while good estimates will enable sharp reconstructions (Fig.~\ref{fig:poseopt}). 

Intuitively, motion blur occurs because the inter-frame motion is not compensated. For conventional cameras, this motion cannot be easily compensated for after the fact. To circumvent this issue, modern smartphones take multiple short exposures and fuse them based on a local motion model such as optical flow~\citeMain{nightsight}. We take this idea to its logical limit with extremely short exposures of single-photon imaging. With quanta radiance fields, \emph{motion blur and pose optimization are tightly interleaved issues}, and the Fourier regularizer introduced above helps us tackle both simultaneously. 

\smallskip\noindent\textbf{Robustness of Fourier Regularizer:} Using the same scene as in Fig.\ref{fig:poseopt}, we progressively add more noise to the initial poses and observe the quality of the final reconstruction in terms of various reconstruction metrics. While the pose noise is sampled from a Gaussian distribution and converted to a pose via the $\mathfrak{s e}(3)$ exponential map, which we then compose with the ground truth poses, we report its effects as mean pose deviations, both in terms of angular deviations and translational displacement, from the ground truth poses for clarity. We vary position and rotation noise independently and report results for a wide range of pose perturbations. 

We present our results in Fig.~\ref{fig:poseopt-metrics}. As a point of comparison, an oracle model trained with fixed ground truth poses would achieve a PSNR of $31.8$, SSIM of $0.86$, and LPIPS of $0.14$. 
Our method achieves similar metrics until very severe noise is added, at which point the reconstruction quality collapses. 
We refer the reader to the supplement for a similar sensitivity analysis that tracks the final pose deviations instead of reconstruction quality.

\begin{figure*}[h]
  \begin{center}
    \includegraphics[width=1.0\textwidth]{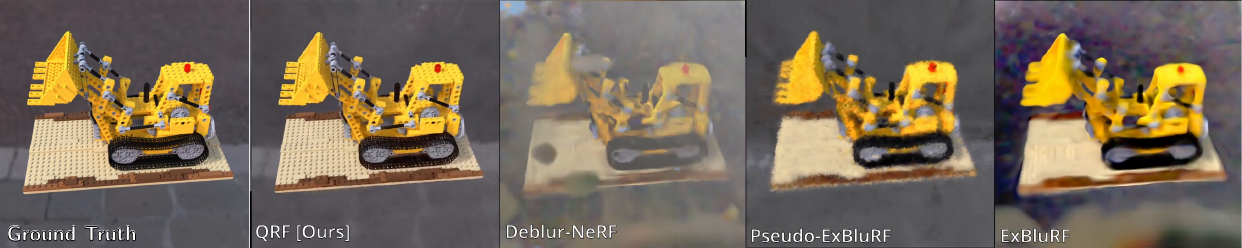}
  \end{center}
  \vspace{-1em}
  \caption{\textbf{Comparison between deblurring methods:} While many motion-aware NeRF methods can lack sharp details, QRFs can successfully recover them.}
  \label{fig:blur-baselines}
\end{figure*}

\smallskip\noindent\textbf{Motion Blur Mitigation in Conventional NeRFs:} Creating radiance fields out of blurry conventional images remains a challenging problem. ExBluRF~\citeMain{lee2023exblurf}, a state-of-the-art approach that addresses motion blur, does so by modeling blur as part of the rendering process. They spawn virtual cameras into the scene and enforce that the average value seen by consecutive cameras within a certain window corresponds to the observed blurry image in the training set. Finally, they promote smooth camera movements by utilizing a B\'ezier curve-based regularizer term on the camera trajectory. 

We re-implement its key features with minor modifications to perform comparisons. Specifically, we replaced their B\'ezier regularizer with our Fourier smoothing regularizer (Eq.~\ref{eq:fft-reg}) as they both constrain and smooth the camera trajectory, with the exception that the proposed Fourier smoothing regularizer scales to hundreds of thousands of virtual cameras. This enables a direct comparison between our method and this modified baseline, which we call pseudo-ExBluRF. Comparisons between pseudo-ExBluRF, the official implementation, and another blur-aware NeRF method can be found in the supplement. 

In Fig.~\ref{fig:blur}, we train the pseudo-ExBluRF method with simulated $50$ fps images from a conventional camera and spawn $20$ additional virtual cameras per training frame (corresponding to a blur kernel of $21$). Camera poses are initialized to their corresponding ground truth pose, or an interpolation of them for the virtual cameras. We train our method on the equivalent dataset which would be captured by a single photon-camera capturing $40$k binary frames per second, and initialize camera poses in the same way, that is, only the cameras corresponding to a $50$ fps conventional camera get initialized with their true poses; every other one is initialized with an interpolated pose. Both methods use the same hyperparameters, and all camera poses are co-optimized with the radiance field. Finally, due to the slow sampling rate of the $50$ fps camera, the low-pass cutoff used in our regularizer is lowered to $25$ Hz. While one might expect a lowpass cutoff, which corresponds to the Nyquist rate of the camera, to not perform any filtering, this is not the case as the regularizer is applied to the virtual cameras as well, which have a combined sampling rate of $50\times21=1050$ Hz.

While pseudo-ExBluRF outperforms the conventional method for the Lego truck (first row of Fig.~\ref{fig:blur}), it fails to recover the mirror sphere, likely due to the specularities and lack of environment map. Better still are the reconstructions with a simulated high-speed camera, which can capture specularities, yet are washed out due to read noise. QRFs outperform these baselines and recover accurate geometry and photometric effects, even in these challenging conditions with rapid motion and high-frequency specular reflectance.

\smallskip\noindent\textbf{Baseline Comparisons:} We now compare our blur-aware pseudo-ExBluRF implementation to the official one and to Deblur-NeRF~\citeSupp{li2022deblurnerf-supp}. The results are shown in Fig.~\ref{fig:blur-baselines}. There are two main differences: first, our implementation uses an FFT-based pose smoothing regularizer while ExBluRF~\citeSupp{lee2023exblurf-supp} uses a per-camera spline, and second, we use an Instant-NGP backbone instead of a Plenoxels~\citeSupp{plenoxels-supp} one.  

The latter is meant to isolate the deblurring components from any variations due to how the scene is represented. By ensuring different models use the same backbone, we can more fairly compare them. The former allows for a more direct comparison to QRFs, as it ensures the camera trajectory is not only piece-wise smooth, but also smooth between frames. 

In practice, we find that ExBluRF~\citeSupp{lee2023exblurf-supp} and our pseudo-ExBluRF perform similarly, with average reconstruction PSNRs of $23.3$ and $23.8$ respectively on the scene shown in Fig.~\ref{fig:blur-baselines}. Yet these models have distinct artifacts, with ExBluRF having a more rounded appearance, with floating blobs, and pseudo-ExBluRF being more cloud-like. We attribute these different artifacts to the different backbone architectures and representations, one being explicit, the other implicit. For a better sense of these artifacts, please see the associated video.

\begin{figure}[h]
  \begin{center}
    \includegraphics[width=1.0\columnwidth]{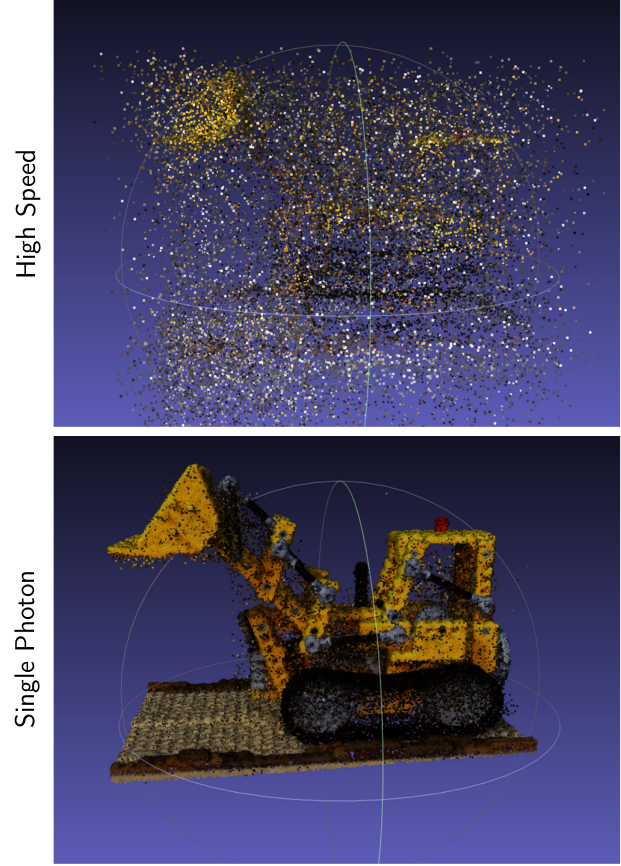}
  \end{center}
  \vspace{-1em}
  \caption{\textbf{Point Clouds Extraction:} We extract point clouds from the conventional NeRF trained on high-speed camera frames and from the QRF model trained on single photon data shown in Fig.~\ref{fig:blur}. Here, the QRF-generated point cloud is much cleaner as the extracted geometry has higher fidelity.}
  \label{fig:pcl}
\end{figure}

\smallskip \noindent \textbf{Point Cloud Generation:}
Using the utilities provided by Nerfstudio~\citeMain{nerfstudio}, we can extract a point cloud from a trained model. In Fig.~\ref{fig:pcl} we show the point cloud extracted from a conventional NeRF model that has been trained using frames from a high-speed camera, as well as the point cloud extracted from a QRF for the scene shown in Fig.~\ref{fig:blur}. QRFs can better recover the Lego truck's geometry, allowing for fewer artifacts when performing view extrapolation.    

\smallskip \noindent \textbf{Challenges of Simulating SPCs:} Currently, simulating long SPC sequences is exorbitantly expensive, primarily due to the high frame rate of SPCs. To speed up data generation, potentially at the cost of interpolation artifacts, we use $8\times$ video interpolation using RIFE~\citeMain{huang2022rife}, allowing a full sequence of $50k-200k$ frames to be rendered in less than a day. In practice, the interpolated frames are very close, thus minimizing interpolation errors.

When artifact-free or non-tonemapped renders are needed, we cannot use this shortcut as many interpolation methods do not work with HDR. For these reasons, the scenes shown in Fig.~\ref{fig:lowflux-hdr} were rendered completely in Blender, taking more than $20$ GPU-days (using an RTX 3090). 
Interpolation was only used to create SPC datasets, not conventional ones, thus the results we show might be slightly worse than if we had artifact-free simulated SPC data.

\smallskip\noindent \textbf{Implementation Details:} We use Nerfstudio's implementation of Instant-NGP~\citeMain{mueller2022instant} as a backbone architecture. We use an Adam optimizer with a learning rate of $0.01$ which decays exponentially to $0.0001$ over the course of $30,000$ training steps, and a batch size of $4096$ rays. Training in this manner takes $\tildeNice 30$ minutes using a single RTX 3090. Unless otherwise noted, we use an ideal low-pass filter with a cut-off of $500$ Hz and a $\lambda$ of $0.1$. All other parameters have been left untouched. Finally, we use Blender~\citeMain{blender} and Eq.~\ref{eq:bernoulli} to simulate binary frames. For more implementation details and code please see the supplement.

\section{Quantitative Evaluations and Additional Experiments}

Thus far, we have shown results only on simulated data as it enables us to emulate different image formation models and to have access to ground truth. However, simulation comes at a high computational cost, with the scene shown in Fig.~\ref{fig:lowflux-hdr} taking more than $20$ GPU days to render fully. 

\begin{figure*}[ht!]
    \centering
    \includegraphics[width=1.0\textwidth]{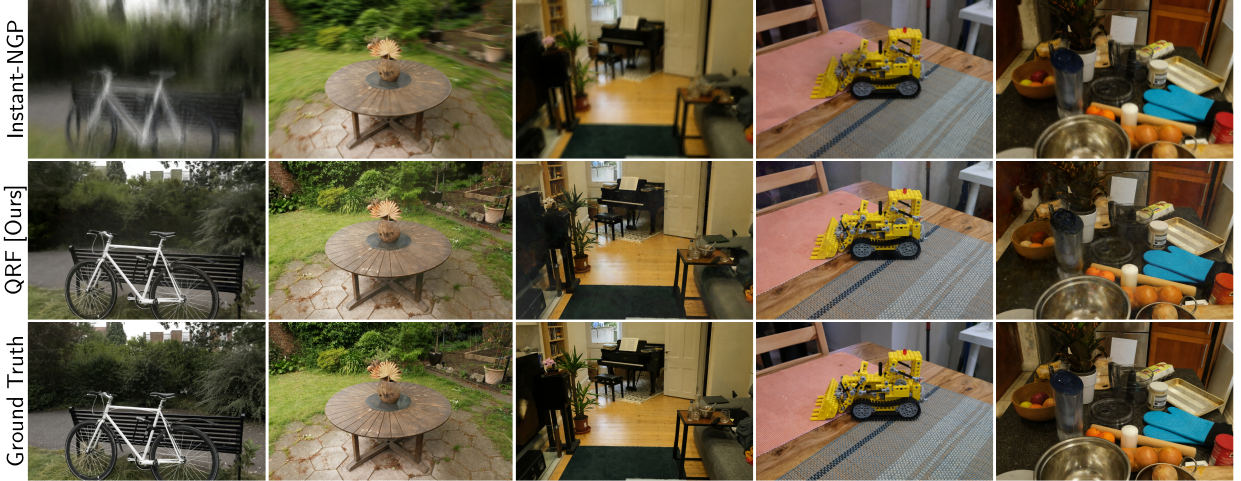}
    \vspace{-1.0em}
    \caption{\textbf{Additional Qualitative Evaluation on Common NeRF Datasets:} Using high-quality pretrained Zip-NeRF models~\protect\citeMain{barron2023zipnerf} we emulate a single photon and conventional camera and render new training sets on which we train a baseline INGP~\protect\citeMain{mueller2022instant} model and a QRF model respectively. While motion blur gets baked into conventional radiance fields, our method using single photon data effectively filters input noise leading to high-quality reconstructions. The type of blur that gets baked in depends on the camera trajectory; we can observe a mostly horizontal blur in the second column, which is due to the orbital trajectory of the camera, while in the third column, the blur is more uniform as the camera trajectory is more complex. The last two columns have noticeably less blur since the scenes are smaller with limited camera movement.  
    \label{fig:mipscenes}}
\end{figure*}

In section~\ref{sec:experiments}, we use a single-photon camera hardware prototype to capture sequences and provide qualitative results on real-world experiments. Here, we instead focus on a quantitative evaluation of our method based on interpolated data from well-known NeRF datasets. This enables us to provide metrics and results on standardized scenes, enabling easier comparisons to other methods. 

\begin{table}[!t]
    \centering
    \caption{\textbf{Average metrics over scenes shown in Fig.~\ref{fig:mipscenes}}}
    \vspace{-0.5em}
    \scalebox{1.1}{
        \begin{tabular}{l|ccc}
            \hline
                           & PSNR$\uparrow$      & SSIM$\uparrow$      & LPIPS$\downarrow$  \\ 
            \hline \hline
            Conventional   & 29.43         & 0.612          & 0.430         \\ 
            Single Photon  & 29.89         & 0.688          & 0.259         \\ 
            \hline 
            Oracle         & 31.62         & 0.880          & 0.075         \\
            \hline 
        \end{tabular}
    }
    \label{tab:metrics}
\end{table}  

\smallskip\noindent\textbf{Dataset Generation:} Since there are no widespread datasets that have been captured with a SPC, we resort to using high-quality pretrained NeRF models as a means to emulate a single-photon capture. Specifically, we use pretrained Zip-NeRF models~\citeMain{barron2023zipnerf} (as released by SMERF~\citeMain{duckworth2023smerf}) and render $2000$ frames along the training trajectory for each scene. From here, we either average neighboring frames in groups of $5$ and apply Gaussian noise to simulate a conventional camera with realistic motion blur, or further interpolate by $16\times$ using RIFE~\citeMain{huang2022rife} and then sample following Eq.~\ref{eq:bernoulli} to simulate our single-photon camera. Following this procedure, we create an oracle (ground-truth) dataset consisting of $2000$ static high-quality frames, a SPC dataset with $32$k binary frames, and a conventional dataset with $400$ RGB frames for each scene.

\smallskip\noindent\textbf{Evaluation Methodology:} To evaluate the upper-bound performance on our new datasets, we train a standard INGP~\citeMain{mueller2022instant} model on the oracle data and report metrics computed between novel views produced by this upper-bound model and the original test set. We further train models on the conventional and single-photon datasets and report metrics in Tab.~\ref{tab:metrics}. 

Although some of these metrics are numerically close, the gap in perceptual quality is rather large, as exemplified qualitatively in Fig.~\ref{fig:mipscenes}. This is especially visible in the bicycle and garden scenes, where the motion blur captured by the conventional camera is baked into the reconstruction, so much so that the motion of the camera during capture can be inferred: the bike scene was recorded with vertically camera movement, whereas the garden scene was captured using a circular trajectory. In contrast, the QRF reconstruction remains clean and blur-free.

\section{Explicit Reconstruction with 3D Gaussian Splatting}
\label{sec:splat}

\begin{figure*}[t]
    \centering
    \includegraphics[width=1.0\linewidth]{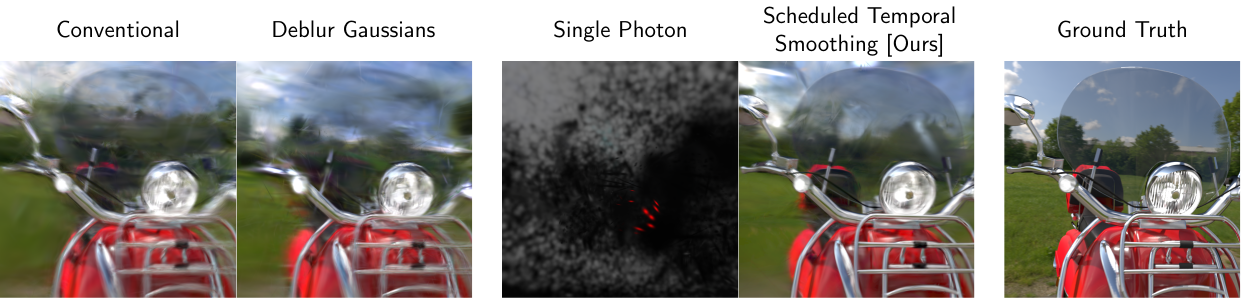}
    \caption{\textbf{Qualitative results from explicit reconstruction:} Traditional 3D Gaussian Splatting~\protect\citeMain{kerbl3Dgaussians} (3DGS) reconstructions suffers with motion-blur data. 
    Deblur Gaussians~\protect\citeMain{deblurring_gaussians} addresses this by modeling per-Gaussian motion offsets using an MLP in the vanilla 3DGS training process. 
    Using purely single photon data does not allow the individual gaussians to converge due to the view-dependent dynamic noise, leading to premature pruning and a failed reconstruction.
    However, we show that temporally smoothing single photon frames allows superior explicit 3D reconstruction. The metrics for the scene are provided in Tab.~\ref{tab:metrics_splatting}.}
    \label{fig:splat}
\end{figure*}

Recently, Kerbl \textit{et al.} introduced 3D Gaussian Splatting (3DGS)~\citeMain{kerbl3Dgaussians}, an explicitly parameterized sparse-view 3D scene reconstruction technique, that offers speed and fidelity advantages over implicit neural radiance field-based methods~\citeMain{rawnerf, mildenhall2020nerf, barron2021mipnerf} when the input frames are clean. 
However, in less ideal conditions, for example, when the capture time is limited or when the subject is poorly illuminated, it is unclear whether this explicit modeling approach achieves better or comparable performance as compared to its implicit counterpart. As shown in Fig.~\ref{fig:splat}, conventional captures can result in blurry reconstructions with inaccurate features and smudged colors. 

In fact, many splatting works pre-process the input sequence or introduce an auxiliary network to improve the fidelity of the image data, rather than model the non-idealities as part of the reconstruction pipeline. 
For instance, HDRSplat~\citeMain{hdrsplat} reconstructs 3D HDR scenes from low-light raw images by first denoising these raw images with a deep neural denoiser (PMRID~\citeMain{pmrid}). 
Similarly, HO-Gaussians~\citeMain{ho_gaussians} and Deblur-Gaussians~\citeMain{deblurring_gaussians} introduce an MLP to process and assist the 3DGS optimization process.

\begin{table}[t]
    \centering
     \caption{\textbf{Average metrics over scene shown in Fig.}~\ref{fig:splat}. }
         \vspace{-0.5em}
    \scalebox{1.1}{
        \begin{tabular}{l|ccc}
            \hline
                              & PSNR$\uparrow$& SSIM$\uparrow$ & LPIPS$\downarrow$  \\ 
            \hline \hline
            Conventional      & 15.15        & 0.692         & 0.216         \\ 
            Deblur Gaussians  & 15.14         & 0.592          & 0.281         \\ 
            \hline 
            TS (Ours) & 16.93         & 0.663         & 0.237 \\
            \hline 
        \end{tabular}
    }
    \label{tab:metrics_splatting}
    \vspace{-0.5em}
\end{table}  

\smallskip\noindent\textbf{Challenges of Explicit Reconstruction with 3DGS:}
When modeling the scene implicitly, the highly stochastic nature of single-photon data is handled by the robustness of modern optimization techniques and by the deep networks that encode the scene. These networks learn the photon detection probability, as a proxy for radiance, which best explains the observations despite their noisy nature. 

On the other hand, when representing the scene as an explicit collection of discrete Gaussian blobs (unlike Plenoxels~\citeMain{plenoxels} that interpolates to produce a continuous representation), what is being learned is not the continuous radiance of a point in space but rather the position, color, and characteristics of each blob that encodes the scene. 
This subtle difference is responsible for splatting's faster training and inference, but also explains why they are not as robust to noise as implicit methods. 
As shown in Fig.~\ref{fig:splat}, training directly on SPAD binary frames leads to an unrecognizable, unstable, and nonconverged point cloud. At its root, the \textit{high view-dynamic noise} present in the binary frames causes aggressive pruning of Gaussians, along with smudging of colors, due to the large gradients that get back-propagated to each blob.

\smallskip\noindent\textbf{Two Step Reconstruction:}
One way to smooth out the large gradients that are responsible for pruning most of the Gaussian blobs is to simply aggregate consecutive binary frames to generate virtual exposures~\cite{Jungerman_2023_ICCV}. While this can reduce the high-view dependent noise, and thus lower premature pruning, it can also lead to unnecessary blur. Similarly to the implicit case, a two-step reconstruction approach would either have to contend with the blur-versus-noise trade-off or have high computational costs if using a more advanced $2$D reconstruction technique. Again, errors would accumulate, as artifacts in the first step would not be corrected subsequently.

\begin{figure*}[t]
    \centering
    \includegraphics[width=1.0\textwidth]{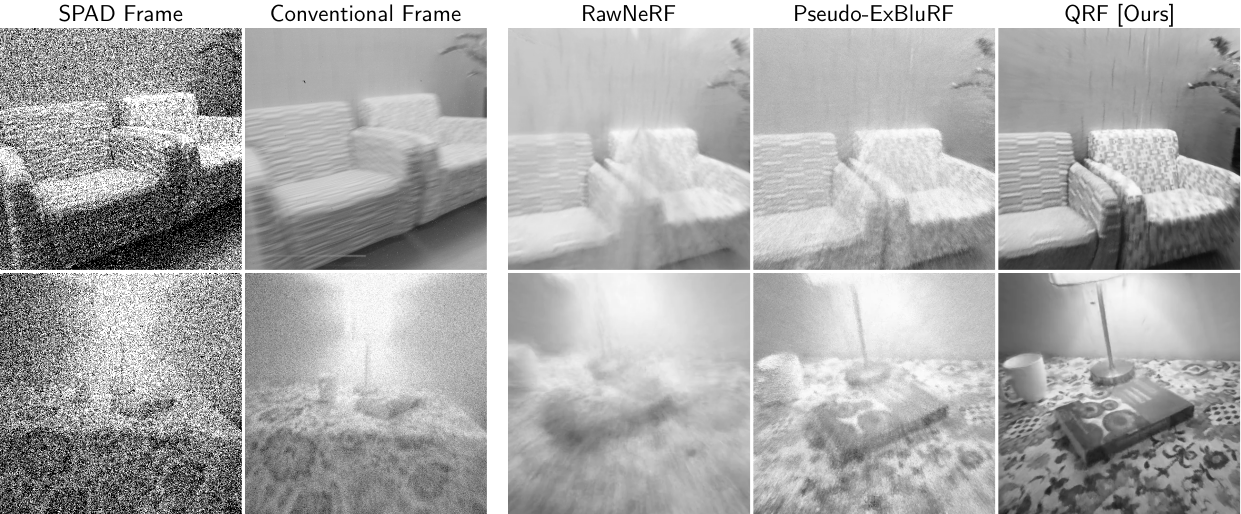}
    \caption{\textbf{Qualitative results on real-world captures:} We capture a room-scale and tabletop scene with a single photon camera in about $8$ seconds and show reconstructions made using emulated raw conventional frames at $30$ fps~\protect\citeMain{rawnerf}, a blur-specific baseline~\protect\citeMain{lee2023exblurf}, and with quanta radiance fields. Overall QRFs exhibit fewer artifacts and better reconstruction quality than baseline methods. Please see the supplemental material for video results of these scenes. \label{fig:realexp}}
\end{figure*}
 
\smallskip\noindent\textbf{Scheduled Temporal Smoothing:} 
To address the problems of large gradients that lead to premature pruning and the overall lack of detail in the final reconstruction, we introduce a two-phase approach: temporal smoothing for initial stabilization and scheduled refinement for detailed recovery.

In the first phase, we temporally smooth the input binary frames by averaging $k$ neighboring frames. This acts as a low-pass filter over a virtual exposure time of $k\tau$, reducing high-frequency noise at the cost of blur. 
The degree of blur is proportional to the number of frames averaged: a higher $k$ results in greater noise reduction but also increased blur. 
However, this smoothing is crucial for stabilizing the initial training phase, allowing the Gaussian blobs to converge to a coherent structure and color representation of the scene, despite not retaining much detail. 
We start with a high value for $k$, which approximates a  conventional cameras (\eg, $25$ fps capture rate for the scene in Fig.~\ref{fig:splat}), and ensures that the blobs stably attain the scene's structure and colors.

Once the blobs have settled into a stable configuration, we transition to the second phase: scheduled refinement. 
Here, we gradually reduce the temporal smoothing effect by decaying the smoothing parameter $k$ from an initial value of multiple thousands down to 1 in a stepwise manner. 
Simultaneously, we decay the learning rates associated with each blob's properties (position, opacity, and features) each time $k$ is decreased. 
This step-wise decay ensures stable training and progressively removes the excess blur introduced in the initial phase. 
By adapting to finer details in the scene without reintroducing instability, this approach achieves a high-quality reconstruction. Fig.~\ref{fig:splat} demonstrates the effectiveness of this method, showing that it outperforms training on conventional frames or SPC frames alone.

It is important to note that this scheduled smoothing and refinement process is performed during the data loading step, making it computationally efficient. 
However, this technique is not necessary for QRFs, as implicit methods like NeRF inherently aggregate information across frames without the risk of pruning. 
In fact, applying temporal smoothing to QRFs can be detrimental, as it introduces unnecessary blur and computational overhead, whereas implicit methods naturally handle noise through their continuous representation.

In summary, while explicit reconstruction methods like Gaussian splatting benefit from temporal smoothing and scheduled refinement to handle noise and stabilize training, implicit methods like NeRF inherently outperform them by better managing high view-dependent noise without the need for such training regimes. More sophisticated pruning regimes have since been proposed~\citeMain{HansonTuPUP3DGS}, and adapting these to work with SPC data is a promising avenue for future research.

\smallskip\noindent\textbf{Quantitative Evaluation:}
We compute mean PSNR, SSIM~\citeMain{ssim-supp} and LPIPS~\citeMain{lpips-supp} using VGG-16~\citeMain{vgg-supp} for Fig.~\ref{fig:splat}'s scene in Tab.~\ref{tab:metrics_splatting} using 500 rendered frames along the same camera trajectory for each method and compare against conventional Gaussian spaltting and a blur-aware method~\citeMain{deblurring_gaussians}.

\section{Real World Experimental Results}
\label{sec:experiments}

We use a SwissSPAD2~\citeMain{SwissSPAD2} single-photon camera, which can be seen in Fig.~\ref{fig:teaser}, to validate our findings using real hardware. This SPAD-based camera is capable of reaching frame rates of $97$ kHz at a resolution of $512 \times 512$. By averaging multiple consecutive binary frames, we can emulate a conventional camera running at any arbitrary slower frame rates, thus enabling direct comparisons between methods. We captured multiple scenes using this camera, two of which are shown in Fig.~\ref{fig:realexp}. These were taken in $\tildeNice 8$ seconds at $40$ kHz, the first was in ambient light while the second was only illuminated using a small nightstand lamp. From this, we emulated raw-intensity $30$ fps conventional frames and reconstructed the scene using multiple techniques. Again, we see higher-quality reconstruction with QRF, with other baselines having washed-out colors and noticeable artifacts. 

We initialize the poses with estimates obtained using COLMAP~\citeMain{colmap} on short virtual exposures. In fact, this initialization is currently the main bottleneck of QRFs and their explicit counterparts; due to the SwissSPAD2's limited resolution, these initial poses are not very precise, and, if the camera motion is too fast, or light levels too low, COLMAP will fail to converge and yield no pose estimates at all. One possible mitigation is to perform frame reconstruction (\eg, QBP~\citeMain{qbp}) before using COLMAP but this comes at a significant computational cost and still relies of structure-from-motion. Alternatively, one could use additional sensors, such as an IMU, or COLMAP-free techniques~\cite{colmap_free_3dgs}, however, these rely on good monocular depth estimation networks instead, which do not yet exist for SPC data. For example failure cases, and more discussion about these limitations, see the supplement. 

Finally, note that these results are in grayscale because our hardware prototype does not include a color filter array; this is a limitation of this specific device and not fundamental to SPADs. Please refer to the supplement for more details about the hardware setup and implementation details, as well as video results.

\section{Discussion and Future Work}
\label{sec:conclusion}

In this work, we show that learning radiance fields at the granularity of single photons has many advantages, including better view extrapolation and reconstruction quality. 
However, many challenges remain. 
Specifically, a key limitation of this approach is that, while camera poses are finetuned during training, a good initial estimate is still required. 
While this limitation is not unique to our method, we highlight it here as currently, this is the key limiting factor for extreme low-light and in-the-wild QRFs. 

Many improvements to 3D reconstruction (NeRFs and 3DGS) have been proposed, and many more will follow. 
New advances, from faster training and inference~\citeMain{garbin2021fastnerf}, to surface rendering approaches~\citeMain{wang2023neus, wolf2024gs2mesh}, or deep priors~\citeMain{yu2020pixelnerf, 3dgs_street_diffusion_prior}, could all benefit from using photons instead of pixels as their basic building blocks. 
These advances are orthogonal directions that will cross-pollinate with the proposed concept and systems of quanta radiance fields.

\bibliographystyleMain{ACM-Reference-Format}
\bibliographyMain{references}

\clearpage
\setcounter{page}{1}

\renewcommand{\figurename}{Supplementary Figure}
\renewcommand{\figurename}{Suppl. Fig.}
 \renewcommand{\thesection}{S.\arabic{section}}
\renewcommand{\theequation}{S\arabic{equation}}
\setcounter{figure}{0}
\setcounter{section}{0}
\setcounter{equation}{0}
\setcounter{page}{1}

\twocolumn[  
    \begin{@twocolumnfalse}
        \begin{center}
        \huge Supplementary Document for\\
        \huge ``Radiance Fields from Photons'' \\
        \vspace{2em}
        \end{center}
     \end{@twocolumnfalse}
]

\section{Simulating Single Photon Cameras}
We use Blender~\citeSupp{blender-supp} to render out ground truth RGB images of a scene for a camera moving along a spline. To save on rendering time, we only render ground truth frames at a simulated $10$ kHz and interpolate these to the speeds achieved by single photon cameras using~\citeSupp{huang2022rife-supp}. This setup enables us to render a sequence in about $8$ hours using a single RTX 3090.

These RGB frames are then sampled using Eq.~\ref{eq:bernoulli} to create binary frames, or Eq.~\ref{eq:camera_model} to create conventional RGB frames with realistic camera blur and noise. \\

\section{Experimental Setup}
Our experimental setup consists of a SwissSPAD2~\citeSupp{SwissSPAD2-supp} single-photon camera along with two Opal Kelly FPGAs, one for each half of the sensor array, and a C-Mount varifocal lens. This sensor is capable of capturing $97$ thousand frames per second at a resolution of $512 \times 512$.

\begin{figure}[ht]
  \begin{center}
    \includegraphics[width=1.0\columnwidth]{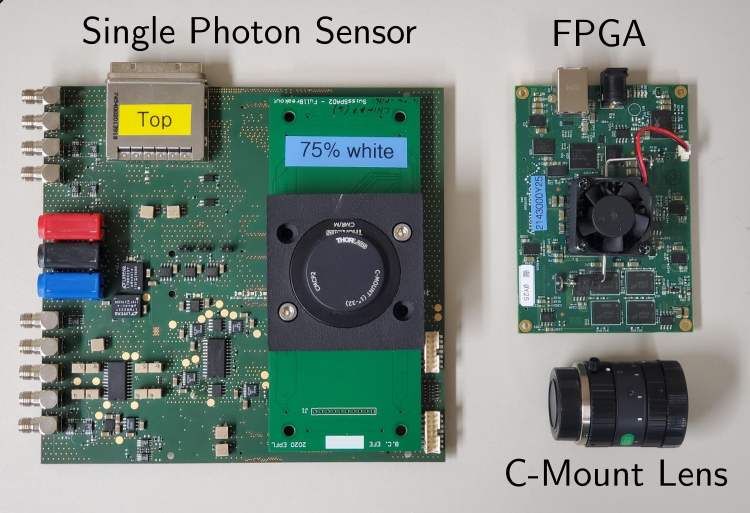}
  \end{center}
  \caption{Hardware Setup}
\end{figure}
 
In practice, we often run this sensor at slightly slower frame rates as it greatly improves the readout reliability. Using it at its maximum framerate limits the total capture time to about two seconds, as the memory buffers fill up faster than the two USB 3.0 interfaces can read off. 

Further, we preprocess the raw data read off this sensor by applying simple filters. First, the two half arrays are read out separately (one FPGA and USB per side), so we must recombine them after the capture is complete. Second, many pixels are dead or always on, so we apply dead-pixel and hot-pixel corrections by simply inpainting these pixels based on their neighbors' values. This fixed pattern correction is typically done by the camera's ISP, before ever reading out the frame.

\section{Additional Pose Regularizer Analysis}
\begin{figure}[h]
  \begin{center}
    \includegraphics[width=1.0\columnwidth]{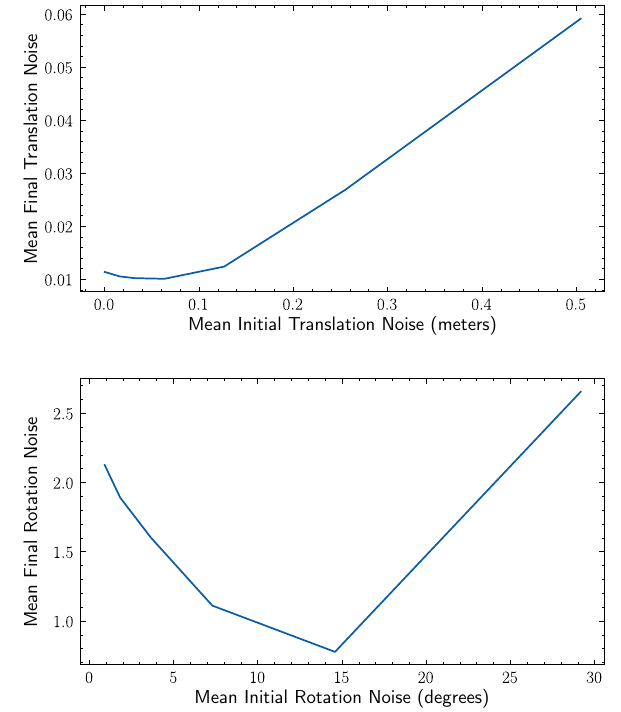}
  \end{center}
  \vspace{-1em}
  \caption{Final pose deviations vs. initial}
  \label{fig:poseopt-supp}
\end{figure}

We complement the analysis shown in Fig.~\ref{fig:poseopt-metrics}, and show the relationship between the mean initial pose deviations and the final deviations after training in Sup.~Fig.~\ref{fig:poseopt-supp}. Notice that while the mean translation error grows with the input translation, it does so at a small rate, with a slope of around $1/10$, while the final mean rotation is nearly constant, fluctuating between $1^\circ - 2.5^\circ$. 

\section{Real World Limitations}
\begin{figure}[t]
  \begin{center}
    \includegraphics[width=1.0\columnwidth]{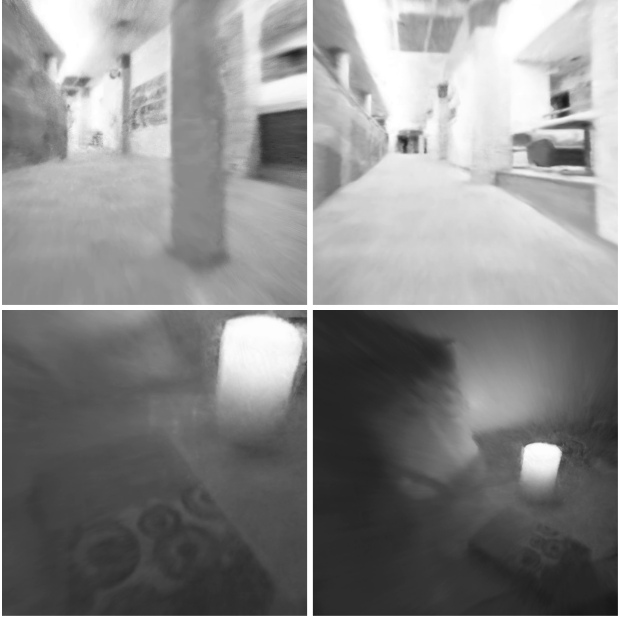}
  \end{center}
  \vspace{-1em}
  \caption{Additional Real World Results}
  \label{fig:real-world-fails}
    \vspace{-1em}
\end{figure}

Fig.~\ref{fig:real-world-fails} shows additional qualitative QRF results on real-world captures for two other scenes. We captured the scene shown on the top row by strapping the SwissSPAD2 on a dolly and running down an indoor corridor, while the second row shows a candlelit tabletop scene that was captured in a handheld manner. The motion in the corridor scene is smooth thanks to the dolly, yet very fast. Although the motion in the tabletop scene is slower, it is only illuminated by a candle. To get a sense of the motion, we refer the reader to the supplemental videos of these scenes, which show the final QRF reconstruction rendered along the COLMAP-estimated acquisition trajectory.

Again, to initialize the camera poses, we locally average binary frames into short virtual exposures and use COLMAP. This works fairly well for easier scenes, but for difficult scenes such as these, the merged frames suffer from blur and noise. Here, COLMAP entirely failed to converge and produce any pose estimates for both of these scenes without careful fine-tuning of hyperparameters. Specifically, to get initial pose estimates for these challenging scenes, we had to tweak COLMAP's SIFT matching arguments. We changed \texttt{max\_error} from $4$ to $20$, \texttt{max\_distance} from $0.69$ to $20.0$, \texttt{confidence} from $0.999$ to $0.0$, \texttt{min\_inlier\_ratio} from $0.25$ to $0.1$, and \texttt{min\_num\_inliers} from $15$ to $4$. These changes allow COLMAP to not prune out subpar matches, further lowering the quality of estimated poses, if any are found at all. 

These poor initial pose estimates currently hinder the reconstruction capabilities of QRFs, leading to at best the blurry results shown in Supp.~Fig.~\ref{fig:real-world-fails} and, at worst, no pose estimates at all. COLMAP-free methods exist, yet we cannot currently apply those techniques to quanta sensor measurements. Adapting these for use here is a promising avenue for future research. 

\section{Quantitative Evaluation and Tonemapping}
Note that since we train QRFs and their splatting counterpart to learn the photon detection probability, we need to first invert the SPAD response function using Eq.~\ref{eq:flux-estimate} and then tonemap the results to sRGB to compute the metrics. 
To convert the linear intensity to sRGB and back, we use the equivalent \href{https://github.com/blender/blender/blob/c29a50a6d9ccee8165711b15820ec9d0312a5f39/source/blender/blenlib/intern/math_color.cc#L423C1-L439C2}{Blender color utilities}, which we port to pytorch.

\section{Explicit Reconstruction: 3D Gaussian Splatting}
3D Gaussian Splatting (3DGS~\citeSupp{kerbl3Dgaussians-supp}) is an explicit 3D reconstruction or differentiable rasterization algorithm that achieves real-time inference speeds at desirable resolutions of 1080p. 
This comes at the cost of much higher representational power offered by deep network-based volumetric radiance optimization algorithms like NeRFs~\citeSupp{mildenhall2020nerf-supp} due to the explicit parameterization and no deep networks in the training procedure. 
We describe the 3DGS algorithm subsequently. 

\subsection{Background}
Given a set of sparse views ($y$) alongside their camera parameters, 3D Gaussian Splatting allows optimizing the point cloud of the scene. 
Each point in the scene is a 3D Gaussian ($G_i$) in world coordinates centered at a unique position vector $\mu_i$ and is further defined by an anisotropic 3D covariance matrix ($\Sigma$)~\citeSupp{world_coords_3dgs-supp}:
\begin{equation}
    G_i(x) = e^{-\frac{1}{2}(\mathbf{x} - \mu_i)^T \Sigma^{-1}_i (\mathbf{x} - \mu_i)}
    \label{eq:g_in_3d_world}
\end{equation}
~\citeSupp{world_coords_3dgs-supp} also demonstrates projection to 2D space given a viewing transformation and derive a projected covariance matrix which is reformulated by Kerbl \textit{et al.}~\citeSupp{kerbl3Dgaussians} to 
\begin{equation}
    \Sigma = \mathcal{R}\mathcal{S}\mathcal{S}^T\mathcal{R}^T
\end{equation}
using scaling ($\mathcal{S}$) and quaternion matrices ($\mathcal{R}$) to satisfy the positive semi-definiteness constraint (Gaussians can not have negative scale or size) during optimization. 

Finally, spherical harmonic (SH) coefficients~\citeSupp{plenoxels-supp} are attached to these $G_i$s to capture view-dependent appearances, and an additional opacity scalar ($\alpha_i$) is learned. 
After radix sorting~\citeSupp{merrill_radix_sort-supp} the Gaussians and projecting them to 2D space, the color of a pixel at location $p$ is decided as follows:
\begin{equation}
    C(p) = \sum_{i=1}^{N} c_i \alpha_i G^{'}_i(p) \prod_{j=1}^{i-1} (1-\alpha_j G^{'}_j(p))
    \label{eq:color_rendering}
\end{equation}
where $N$ is the total gaussians, $c_i$ is the color and $G^{'}_i(p)$ is the $i^{th}$ 3D gaussian projected to the 2D image space.

\smallskip\noindent\textbf{Loss:}
After rendering all colors/pixels for a view, an $\mathcal{L}_1$ and $D-SSIM$ loss is computed between the rendered and the ground truth image as follows:
\begin{equation}
    \mathcal{L}_{3dgs} = \lambda_{\text{ddsim}}\| \hat{y} - y \|_1   + (1 - \lambda_{\text{ddsim}}) \text{DDSIM}(\hat{y}, y)
    \label{eq:loss3dgs}
\end{equation}
We use the default $\lambda_{\text{ddsim}} = 0.2$ for all our experiments. 

\subsection{Add Points Algorithm}
\label{subsec:add-points}
Initializing a higher number of Gaussians per scene or starting with a denser point cloud is believed to result in superior reconstructions based on the argument that more scene details can be encoded. Extrapolating the same argument, one can theoretically populate the scene with infinite Gaussians and model the volume precisely at any granularity. 
Following a similar idea, Deblur Gaussians~\citeSupp{deblurring_gaussians-supp} introduced the Add-Points algorithm to make the point cloud dense during optimization since a dense point cloud or random initialization is not guaranteed to be dense. 
The algorithm simply initializes $N_{extra}$ Gaussians and spreads them Uniformly throughout the scene. 
Then it uses KNN to find neighbors ($4$ in our case) to interpolate the scale, rotation and color properties for optimization.
This interpolation is required to not get the points pruned in the next iteration (since they are likely to receive large gradients if all parameters are randomly initialized).

\subsection{Implementation Details}
We use 3DGS's~\citeSupp{kerbl3Dgaussians-supp} implementation that uses PyTorch~\citeSupp{pytorch-supp}, CUDA kernels for rasterization~\citeSupp{rasterizer_kopanas-supp} and NVIDIA CUB fast radix sort~\citeSupp{merrill_radix_sort-supp} for all our experiments. 
Additionally, we used random point cloud initializations instead of COLMAP~\citeSupp{colmap-supp} initializations, for all experiments since it worked better overall.

The conventional camera captures are trained with default parameters except for Gaussian densification (20k) and total number of iterations (50k). 
This allows for encoding the scene with a much larger number of Gaussians which is beneficial for reconstruction~\citeSupp{kerbl3Dgaussians-supp, deblurring_gaussians-supp, ho_gaussians-supp}. 
The position learning rate (LR) is initialized at $1.6 \times 10^{-4}$ and is exponentially decayed to $1.6 \times 10^{-6}$. All other LRs are kept constant. Note that Adam~\citeSupp{Kingma2014AdamAM-supp} is used to optimize.

For our temporal smoothing single photon solution, we use the same hyper-parameters as above however we train with 25 fps (highest smoothing) for 20k iterations and then decay all LRs (except position which has it's own exponential decay scheduler) to 50\%. 
Subsequently, we train for another 15k iterations with 50fps smoothing and decay all other LRs to 10\% of their previous value, then 10k iterations with 100fps smoothing and 5\% of previous LRs step, then another 5k iterations with 200fps and 2\% LRs and finally 2.5k iterations with 1000fps smoothing (less smoothing).
This progressively lesser smoothing allows for adding more details (less blurring) to the scene at the cost of adding noise. 
However, the initial iterations help stabilize the point cloud (position, rotation and scale at least) and the decayed LRs don't allow the Gaussians to have large gradients, which could lead to harmful pruning.

For our binary or single photon-only training schedule we have to employ the Add-Points algorithm (\ref{subsec:add-points}) from~\citeSupp{deblurring_gaussians-supp} to counter the aggressive pruning observed from the initial iterations. 
Additionally, we lower the learning rates significantly for the same reason. 
We initialize the position LR to 50\% of the default, scaling LR at $1e^-3$, opacity significantly lower at $5e-3$ to counter the dynamic large view-dependent noise patterns, feature LR at $5e^-4$, increase the spherical harmonics degree to the maximum value (3) to have the highest representation power for colors and use the Add-Points algorithm at a sample interval of 1.5k iterations.
Additionally, we do not use the D-SSIM loss for binary frame training.

We use Deblur Gaussians~\citeSupp{deblurring_gaussians-supp} out of the box with the provided synthetic camera motion parameters to train on the (simulated) conventional camera dataset.

\newpage
\bibliographystyleSupp{ACM-Reference-Format}
\bibliographySupp{references-supp}

\end{document}